\documentclass{article}

\usepackage[final]{neurips_2022}

\usepackage[utf8]{inputenc} % allow utf-8 input
\usepackage[T1]{fontenc}    % use 8-bit T1 fonts
\usepackage[colorlinks=true, citecolor=blue, linkcolor=black]{hyperref}      % hyperlinks
\usepackage{url}            % simple URL typesetting
\usepackage{booktabs}       % professional-quality tables
\usepackage{amsfonts}       % blackboard math symbols
\usepackage{nicefrac}       % compact symbols for 1/2, etc.
\usepackage{microtype}      % microtypography
\usepackage{xcolor}         % colors
\usepackage{graphicx}
\usepackage{multirow}
\usepackage{cleveref}
\usepackage{algorithm}
\usepackage{algorithmic}
\usepackage{adjustbox}
\usepackage{wrapfig}
\usepackage{subfigure}
\title{Efficient Dataset Distillation \\ using Random Feature Approximation}

\author{%
  Noel Loo,~~Ramin Hasani,~~Alexander Amini,~~Daniela Rus \\%\thanks{Use footnote for providing further informationabout author (webpage, alternative address)---} \\
  Computer Science and Artificial Intelligence Lab (CSAIL)\\
  Massachusetts Institute of Technology (MIT)\\
  \texttt{\{loo, rhasani, amini, rus\} @mit.edu} \\
}

\begin{document}

\maketitle

\begin{abstract}
Dataset distillation compresses large datasets into smaller synthetic coresets which retain performance with the aim of reducing the storage and computational burden of processing the entire dataset. Today's best-performing algorithm, \textit{Kernel Inducing Points} (KIP), which makes use of the correspondence between infinite-width neural networks and kernel-ridge regression, is prohibitively slow due to the exact computation of the neural tangent kernel matrix, scaling $O(|S|^2)$, with $|S|$ being the coreset size. To improve this, we propose a novel algorithm that uses a random feature approximation (RFA) of the Neural Network Gaussian Process (NNGP) kernel, which reduces the kernel matrix computation to $O(|S|)$.  Our algorithm provides at least a 100-fold speedup over KIP and can run on a single GPU. Our new method, termed an RFA Distillation (RFAD), performs competitively with KIP and other dataset condensation algorithms in accuracy over a range of large-scale datasets, both in kernel regression and finite-width network training. We demonstrate the effectiveness of our approach on tasks involving model interpretability and privacy preservation.\footnote{Code is available at \texttt{https://github.com/yolky/RFAD}}
\end{abstract}

%%%%%%%%%%%%%%%%%%%%%%%%%%%%%%%%%%%%%%%%%%%%%%%%%%%%%%%%%%%%

\begin{wrapfigure}[15]{r}{0.5\textwidth}
\vspace{-14mm}
\begin{center}
% \centerline{\includegraphics[width=\columnwidth]{figures/interpretability/incorrect_0.PNG}}
\includegraphics[scale=1.0]{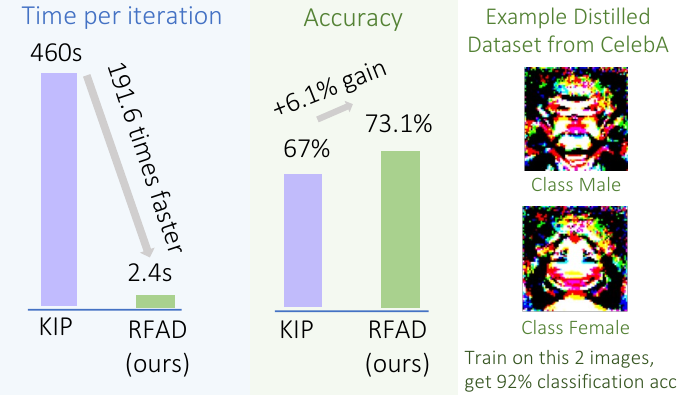}
\vspace{-4mm}
\caption{RFAD provides over 100-fold speedup over the state-of-the-art algorithm Kernel-Inducing Points (KIP) \citep{KIP1}, while exceeding its performance on CIFAR-10. (right) Example distilled synthetic sets one image per class}
\label{icml-historical1}
\end{center}
\end{wrapfigure}
\section{Introduction}
Coreset algorithms aim to summarize large datasets into significantly smaller datasets that still accurately represent the full dataset on downstream tasks \citep{coresets_intro}. There are myriad applications of these smaller datasets including speeding up model training \citep{craig}, reducing catastrophic forgetting \citep{gradient_coresets_continual,icarl,borsos2020coresets}, and enhancing interpretability \citep{prototypes_and_criticisms,prototype_interp2}. While most coreset selection techniques aim to select representative data points from the dataset, recent work has looked at generating synthetic data points instead, a process known as dataset distillation \citep{wang2018dataset,bohdal2020flexible_distillation,sucholutsky2019soft_distillation,zhao2021DC,zhao2021dsa,KIP2}. These synthetic datasets have the benefit of using continuous gradient-based optimization techniques rather than combinatorial methods and are not limited to the set of images and labels given by the dataset, providing added flexibility and performance. 

A large variety of applications benefit from obtaining an efficient dataset distillation algorithm. For instance, Kernel methods \citep{vinyals2016matching,metriclearning,snell2017prototypical,ghorbani2020neural,refinetti2021classifying} usually demand a large support set in order to generate good prediction performance at inference. This can be facilitated by an efficient dataset distillation pipeline. Moreover, distilling a synthetic version of sensitive data helps preserve privacy; a support set can be provided to an end-user for the downstream applications without disclosure of data. Lastly, for resource-hungry applications such as continual learning \citep{borsos2020coresets}, neural architecture search \citep{shleifer2019using} and automated machine learning \citep{hutter2019automated}, generation of a support-set on which we can fit models efficiently is very helpful.

Recently, a dataset distillation method called Kernel-Inducing Points (KIP) \citep{KIP1,KIP2} showed great performance in neural network classification tasks. KIP uses Neural Tangent Kernel (NTK) ridge-regression to \textit{exactly} compute the output states of an infinite-width neural network trained on the support set. Although the method established the state-of-the-art for dataset distillation in terms of accuracy, the computational complexity of KIP is very high due to the exact calculation of the NTK. The algorithm, therefore, has limited applicability.

In this paper, we build on the prior work on KIP and develop a new algorithm for dataset distillation called RFAD, which has similar accuracy and significantly better performance than KIP. The key insight is to introduce a new kernel inducing point method that improves complexity from $O(|S|^2)$ (where $|S|$ is the support-set size) to $O(|S|)$. To this end, we make three \textbf{major contributions}:

\textbf{I.} We develop RFAD, a fast, accurate, and scalable algorithm for dataset distillation in neural network classification tasks.

\textbf{II.} We improve the time performance of KIP \citep{KIP1,KIP2} by over two orders of magnitude while retaining or improving its accuracy. This speedup comes from leveraging a random-feature approximation of the Neural Network Gaussian Process (NNGP) kernel by instantiating random neural networks. 

\textbf{III.} We show the effectiveness of RFAD in efficient dataset distillation tasks, enhancing model interpretability and privacy preservation.

\section{Background and Related Work}
\label{sec:background}

\textbf{Coresets and Dataset Distillation.} Coresets are a subset of data that ensure models trained on them show competitive performance compared to models trained directly on data. Standard coreset selection algorithms use importance sampling to find coresets \citep{har2004coresets,lucic2017training,cohen2017input}. Besides random selection methods, inspired by catastrophic forgetting \citep{toneva2019empirical} and mean-matching (adding samples to the coreset to match the mean of the original dataset) \citep{chen2010super,icarl,castro2018end,belouadah2020scail, input_space_versus_feature_space}, new algorithms have been introduced. An overview of how coresets work on point approximation is provided in \citet{phillips2016coresets}

More recently, aligned with coreset selection methods, new algorithms have been developed to distill a synthetic dataset from a given dataset, such that fitting to this synthetic set provides performance on par with training on the original dataset \citep{wang2018dataset}. To this end, these dataset condensation (or distillation) algorithms use gradient matching \citep{zhao2021DC,maclaurin2015gradient,lorraine2019optimizing}, utilize differentiable siamese augmentation \citep{zhao2021dsa}, and matching distributions \citep{zhao2021dataset}. Dataset distillation has also been applied to the labels rather than images \citep{bohdal2020flexible_distillation}. Recently, a novel algorithm called Kernel Inducing Points (KIP) \citep{KIP1,KIP2} has been introduced that performs very well on distilling synthetic sets by using neural tangent kernel ridge-regression (KRR). KIP, similar to other algorithms, is computationally expensive. Here, we propose a new method to significantly improve its complexity.

\textbf{Infinite Width Neural Networks.} Single-layer infinite-width randomly initialized neural networks correspond to Gaussian Processes \citep{Neal_old_BNN}, allowing for closed-form exact training of Bayesian neural networks for regression. Recently, this has been extended to deep fully-connected networks \citep{GP_in_wide_deep1, GP_in_wide_deep2}, convolutional networks \citep{GP_convolution1, GP_convolution2}, attention-based networks \citep{attention_GP}, and even to arbitrary neural architectures \citep{tensor_programs_are_GPs}, with the corresponding GP kernel being the NNGP Kernel. Likewise, for infinite-width neural networks trained with gradient descent, the training process simplifies dramatically, corresponding to kernel ridge regression when trained with MSE loss with the corresponding kernel being the Neural Tangent Kernel (NTK) \citep{Jacot2018ntk, Arora_Exact_NTK_calc, loo2022evolution}. These two kernels are closely related, as the NNGP kernel forms the leading term of the NTK kernel, representing the effect of the final layer weights. Calculation of kernel entries typically scales with $O(HWD)$ for conv nets, with $H, W$ being the image height and width, $D$ the network depth, and $O(H^2W^2D)$ for architectures with global average pooling \citep{Arora_Exact_NTK_calc}. This, combined with the necessity of computing and inverting the $N \times N$ kernel matrix for kernel ridge regression, typically make these methods intractable for large datasets \citep{snelson2006sparse,titsias2009variational}.

\textbf{Random Feature methods.} Every kernel corresponds to a dot product for some feature map: $k(x,x') = \phi(x)^T\phi(x')$. Random feature methods aim to approximate the feature vector with a finite-dimensional random feature vector, the most notable example being Random Fourier Features \citep{rahimi_rff}. Typically, this limits the rank of the kernel matrix, enabling faster matrix inversion and allowing for scaling kernel methods to large datasets. Recently, these random feature methods have been used to speed up NTKs and NNGPs \citep{NTK_features_via_sketching,fast_finite_NTK, GP_convolution1} at inference or for neural architecture search \citep{NNGP_architecture_search}. In this work, we focus on the NNGP approximation described in \citet{GP_convolution1}, as it only requires network forward passes and is model agnostic, allowing for flexible usage across different architectures without more complex machinery needed to calculate the approximation, unlike those found in \citet{NTK_features_via_sketching, fast_finite_NTK}.

\section{Algorithm Setup and Design}
\label{sec:method}
In this section, we first provide a high-level background on the KIP algorithm. We then sequentially outline our modifications leading to the RFAD algorithm.

\subsection{KIP Revisit} 
The Kernel-Inducing Point algorithm \citep{KIP1, KIP2}, or KIP, is a dataset distillation technique that uses the NTK kernel ridge-regression correspondence to compute exactly the outputs of an infinite-width neural network trained on the support set, bypassing the need to ever compute gradients or back-propagate on any finite network. Let $X_T, y_T$ correspond to the images and one-hot vector labels on the training dataset and let $X_S, y_S$ be the corresponding images and labels for the support set, which we aim to optimize. We have the outputs of a trained neural network as $f(X_T) = K_{TS}(K_{SS} + \lambda I )^{-1}y_S$, with $K$ being the kernel matrices calculated using the NTK kernel, with $T\times S$ or $S \times S$ entries, for $K_{TS}$ and $K_{SS}$, respectively. $\lambda$ is a small regularization parameter. KIP then optimizes $L_{MSE} = ||y_T - f(X_T)||_2^2$ directly. The key bottleneck is the computation of these kernel matrices, requiring $O(TS \cdot HWD)$ time and memory, necessitating the use of hundreds of GPUs working in parallel. Additionally, the use of the MSE loss is suboptimal. 

\subsection{Replacing the NTK Kernel with an NNGP Kernel}
We first replace the NTK used in the kernel regression of KIP with an NNGP kernel. While this change alone would yield a speed up, as the NNGP kernel is less computationally intensive to compute \citep{neural_tangents}, we primarily do this because the NNGP kernel admits a simple random feature approximation, with advantages described later in this section. We first justify the appropriateness of this modification.

Firstly, we denote that in the computation of NTK ($\Theta$) and NNGP ($K$) forms the leading term, as shown in Table 1 in Appendix D of \citep{neural_tangents} which outlines the NTK and NNGP kernel computation rules for various layers of a neural network. For fully connected (FC) layers, which is the typical final layer in neural network architectures, the remaining terms are suppressed by a matrix of expected derivatives with respect to activations, $\dot{K}$, as observed by the recursion yielded from the computation of the NTK for an FC network \citep{neural_tangents}: $\Theta^l = K^l +  \dot{K}^l \odot \Theta ^ {l-1}$. For ReLU activations, the entries in this derivative matrix are upper bounded by $1$, so the remaining terms must have a decaying contribution. We verify that our algorithms still provide good performance under the NTK and for finite networks trained with gradient descent, justifying this approximation.

\subsection{Replacing NNGP with an Empirical NNGP}
When we sample from a Gaussian process $f\sim \mathcal{GP}(0, K)$, it suggests a natural finite feature map corresponding to scaled draws from the GP: $\hat{\phi}(x) = \frac{1}{\sqrt N}[f_1(x), ... , f_N(x)]^T$. For most GPs, this insight is not relevant, as sampling from a GP typically requires a Cholesky decomposition of the kernel matrix, requiring its computation in the first place \citep{gp_bible}. However, for NNGP we can generate approximate samples of $f$ by instantiating random neural networks, $f_i(x) = f_{\theta_i}(x), \theta_i \sim p(\theta)$, for some initialization distribution $p(\theta)$.
\begin{algorithm}[t!]
\caption{Dataset distillation with NNGP random features}\label{alg:RFAD}
\begin{algorithmic}
\REQUIRE Training set and labels $X_T, y_T$,~ Randomly initialized coreset and labels $X_S, y_S$,~ 
Random network count $N$,~
Random network output dimension $M$,~ 
Batch size $|B|$,~ 
Random network initialization distribution, $p(\theta)$,~
Regularization coefficient, $\lambda$,~
Learning rate $\eta$,~
\WHILE{loss not converged}
\STATE{Sample batch from the training set $X_B, y_B \sim p(X_T, y_T)$}
\STATE{Sample $N$ random networks each with output dimension $M$ from $p(\theta)$: $\theta_1, ...\theta_N \sim p(\theta)$}
\STATE{Compute random features for batch with random nets: \\${\scriptstyle \quad\quad\quad\hat{\Phi}(X_B) \gets \frac{1}{\sqrt{NM}}[f_{\theta_1}(X_B), ... , f_{\theta_N}(X_B)]^T \in \mathbb{R}^{|NM|\times|B|}}$}
\STATE{Compute random features for support set with random nets: \\${\scriptstyle \quad\quad\quad\hat{\Phi}(X_S) \gets \frac{1}{\sqrt{NM}}[f_{\theta_1}(X_S), ... , f_{\theta_N}(X_S)]^T\in \mathbb{R}^{|NM|\times|S|}}$}
\STATE{Compute kernel matrices: $\hat{K}_{BS} \gets \hat{\Phi}(X_B)^T \hat{\Phi}(X_S)$}
\STATE{$\hat{K}_{SS} \gets \hat{\Phi}(X_S)^T \hat{\Phi}(X_S)$}
\STATE{Calculate trained network output on batch: $\hat{y}_B \gets \hat{K}_{BS}(\hat{K}_{SS} + \lambda I_{|S|})^{-1}y_S$}
\STATE{Calculate loss: $\mathcal{L} = \mathcal{L}(y_B, \hat{y}_B)$}
\STATE{Update coreset: $X_S \gets X_S - \eta\frac{\partial \mathcal{L}}{\partial X_S}, y_S \gets y_S - \eta\frac{\partial \mathcal{L}}{\partial y_S}$}
\ENDWHILE
\end{algorithmic}
\end{algorithm}
Moreover, with a given neural network, we can define $f_i$ to be a vector of dimension $M$ by having a network with multiple output heads, meaning that with $N$ networks, we have $N~M$ features. For our purposes, we typically have N = 8, M = 4096, giving 32768 total features. For the convolutional architectures we consider, this corresponds to $C = 256$ convolutional channels per layer. Even with this relatively large number of features, we still see a significant computation speedup over exact calculation.

To sample $f\sim \mathcal{GP}(0, K)$, we would have to instantiate random \textit{infinite} width neural nets, whereas, in practice, we can only sample finite ones. This discrepancy incurs an $O(1/C)$ bias to our kernel matrix entries, with $C$ being the width-relevant parameter (i.e., convolutional channels) \citep{yaida_man}. However, we have a $O(1/(NC))$ variance of the mean of the random features \citep{compositional_kernels}, meaning that in practice, the variance dominates the computation over bias. This has been noted empirically in \citet{GP_convolution1}, and we verify that the finite-width bias does not significantly affect performance in \cref{app:sec_empirical_inference}, showing that we can achieve reasonable performance with as little as \textit{one} convolution channel.

The time cost of computing these random features is linear in the training set and coreset size, $|T|, |S|$. With the relatively low cost of matrix multiplication, this results in the construction of the kernel matrices $K_{TS}$ and $K_{SS}$ having $O(|T|+|S|)$ and $O(|S|)$, time complexity, respectively, as opposed to $O(|T||S|)$ and $O(|S|^2)$ with KIP. Noting that the cost of matrix inversion is relatively small compared to random feature construction, our total runtime is reduced to \textbf{linear} in the coreset size. We empirically verify this linear time complexity in section 4.1 and additionally provide a more detailed discussion in \cref{app:more_time_complexity}.

\subsection{Loss Function in dataset distillation} 
We denoted earlier that $L_{MSE}$ is not well suited for dataset distillation settings. In particular, there are two key problems: 

\textbf{Over-influence of already correctly classified data points.} Consider two-way classification, with the label $1$ corresponding to the positive class and $-1$ corresponding to the negative class. Let $x_1$ and $x_2$ be items in the training set whose labels are both $1$. Let $f_{\tiny{\textrm{KRR}}}(x) = K_{x,S}(K_{SS} + \lambda I )^{-1}y_S$ be the KRR output on $x$ given our support set $X_S$. If $f_{\tiny{\textrm{KRR}}}(x_1) = 5$ and $f_{\tiny{\textrm{KRR}}}(x_2) = -1$, then the resulting MSE error on $x_1$ and $x_2$ would be $16$ and $4$, respectively. Notably, $x_1$ incurs a larger loss and results in a larger gradient on $X_S$ than $x_2$, despite being correctly classified and $x_2$ being incorrectly classified. In the heavily constrained dataset distillation setting, fitting both data points simultaneously is not possible, leading to underfitting of the data in terms of classification in order to better fit already-correctly labeled data points in terms of regression. 

\textbf{Unclear probabilistic interpretation of MSE for classification.} This prevents regression from being used directly in calibration-sensitive environment, necessitating the use of transformation functions in tasks such as GP classification \citep{GPs_for_classification, dirichlet_GP_classificaiton_why}.

Based on these two issues, we adopt : 
Platt scaling \citep{platt_scaling}, by applying a cross entropy loss to the labels instead of an MSE one: $\mathcal{L}_{\tiny{\textrm{platt}}} = \textrm{x-entropy}(y_T, f(X_T)/\tau)$,
where $\tau$ is a positive learned temperature scaling parameter. Unlike typical Platt scaling, we learn $\tau$ jointly with our support set instead of post-hoc tuning on a separate validation set. $f(X_T)$ is still calculated using the same KRR formula. Accordingly, this corresponds to training a network using MSE loss, but at inference, scaling the outputs by $\tau^{-1}$ and applying a softmax to get a categorical distribution. Unlike typical GP classification, we ignore the variance of our predictions, taking only the mean instead.

The combination of these three changes, namely, using the NNGP kernel instead of NTK, applying a random-feature approximation of NNGP, and Platt-scaling result in our RFAD algorithm, which is given in \cref{alg:RFAD}.

\section{Experiments with RFAD}
\label{sec:results}
Here, we perform experiments to evaluate the performance of RFAD in dataset distillation tasks.

\textbf{Benchmarks.} We applied our algorithm to five datasets: MNIST, FashionMNIST, SVHN, CIFAR-10 and CIFAR-100 \citep{lecun2010mnist, xiao2017fashion, netzer2011reading_svhn, krizhevsky2009learning_cifar}, distilling the datasets to coresets with 1, 10 or 50 images per class.

\begin{table*}[t]
% \onecolumn
\centering
\small
\caption{Kernel distillation results on five datasets with varying support set sizes. \textbf{Bolded} numbers indicate the best performance with fixed labels, and \underline{underlined} numbers indicate the best performance with learned labels. Note that DC and DSA use fixed labels. %RFAD outperforms all other algorithms on nearly all datasets when using fixed labels; however, with learned labels, KIP pulls ahead, with RFAD only seeing a minor performance increase with label learning. 
(n = 4)}
\begin{adjustbox}{width=1\textwidth}
\begin{tabular}{cccccccc}\toprule
                              &         &            \multicolumn{4}{c}{Fixed Labels} & \multicolumn{2}{c}{Learned Labels} \\\midrule
                              & Img/Cls & DC           & DSA          & KIP             & RFAD (ours)           & KIP                  & RFAD (ours)        \\\midrule
\multirow{3}{*}{MNIST}        & 1       & $91.7\pm0.5$ & $88.7\pm0.6$ & $95.2\pm0.2$    & $\mathbf{96.7\pm0.2}$   & $\underline{97.3\pm0.1}$         &    $\underline{97.2\pm0.2}$    \\
                              & 10      & $97.4\pm0.2$ & $97.8\pm0.1$ & $98.4\pm0.0$    & $\mathbf{99.0\pm0.1}$   & $\underline{99.1\pm0.1}$         & $\underline{99.1\pm0.0}$  \\
                              & 50      & $98.8\pm0.1$ & $\mathbf{99.2\pm0.1}$ & $99.1\pm0.0$    & $99.1\pm0.0$   & $\underline{99.4\pm0.1}$         &$99.1\pm0.0$ \\\midrule
\multirow{3}{*}{Fashion-MNIST} & 1       & $70.5\pm0.6$ & $70.6\pm0.6$ & $78.9\pm0.2$    & $\mathbf{81.6\pm0.6}$   & $82.9\pm0.2$         &$\underline{84.6\pm0.2}$ \\
                              & 10      & $82.3\pm0.4$ & $84.6\pm0.3$ & $87.6\pm0.1$    & $\mathbf{90.0\pm0.1}$   & $\underline{91.0\pm0.1}$         &$90.3\pm0.2$ \\
                              & 50      & $83.6\pm0.4$ & $88.7\pm0.2$ & $90.0\pm0.1$    & $\mathbf{91.3\pm0.1}$   & $\underline{92.4\pm0.1}$         &$91.4\pm0.1$ \\\midrule
\multirow{3}{*}{SVHN}         & 1       & $31.2\pm1.4$ & $27.5\pm1.4$ & $48.1\pm0.7$    & $\mathbf{51.4\pm1.3}$   & $\underline{64.3\pm0.4}$         & $57.4\pm.8$\\
                              & 10      & $76.1\pm0.6$ & $\mathbf{79.2\pm0.5}$ & $75.8\pm0.1$    & $77.2\pm0.3$   & $\underline{81.1\pm0.5}$         &$78.2\pm0.5$ \\
                              & 50      & $82.3\pm0.3$ & $\mathbf{84.4\pm0.4}$ & $81.3\pm0.2$    & $81.8\pm0.2$   & $84.3\pm0.1$         &$82.4\pm0.1$ \\\midrule
\multirow{3}{*}{CIFAR-10}     & 1       & $28.3\pm0.5$ & $28.8\pm0.7$ & $59.1\pm0.4$    & $\mathbf{61.1\pm0.7}$   & $\underline{64.7\pm0.2}$         &$61.4\pm0.8$ \\
                              & 10      & $44.9\pm0.5$ & $52.1\pm0.5$ & $67.0\pm0.4$    & $\mathbf{73.1\pm0.1}$   & $\underline{75.6\pm0.2}$         &$73.7\pm0.2$ \\
                              & 50      & $53.9\pm0.5$ & $60.6\pm0.5$ & $71.7\pm0.2$    & $\mathbf{76.1\pm0.3}$   & $\underline{80.6\pm0.1}$         &$76.6\pm0.3$\\\midrule
\multirow{2}{*}{CIFAR-100}    & 1       & $12.8\pm0.3$ & $13.9\pm0.3$ & $31.8\pm0.3$    & $\mathbf{36.0\pm0.4}$   & $34.9\pm0.1$         &$\underline{44.1\pm0.1}$\\
                              & 10      & $25.2\pm0.3$ & $32.3\pm0.3$ & $\mathbf{46.0\pm0.2}$    & $44.9\pm0.2$   & $\underline{49.5\pm0.3}$         &$46.8\pm0.2$ \\\bottomrule
\end{tabular}
\end{adjustbox}
\label{tab:accuracy_plot}
\end{table*}

\begin{table}[t]
\centering
\caption{Performance of finite networks trained with gradient descent on DC/DSA, KIP, and RFAD distilled images. %RFAD has the best performing finite-network performance on most datasets, which we attribute to centering and label scaling for finite networks. 
* denotes the result was obtained using learned labels. (n = 12)}
\resizebox{0.6\columnwidth}{!}{
\begin{tabular}{ccccc} 
\toprule
                                                                         & Img/Cls & DC/DSA                & KIP to NN              & RFAD to NN                 \\ 
\midrule
\multirow{3}{*}{MNIST}                                                   & 1       & $91.7\pm0.5$          & $90.1\pm0.1$           & $\mathbf{94.4\pm1.5}^*$       \\
                                                                         & 10      & $97.8\pm0.1$          & $97.5\pm0.0$           & $\mathbf{98.5\pm0.1}^*$      \\
                                                                         & 50      & $\mathbf{99.2\pm0.1}$ & $98.3\pm0.1$           & $98.8\pm0.1$  \\ 
\midrule
\multirow{3}{*}{\begin{tabular}[c]{@{}c@{}}Fashion-\\MNIST\end{tabular}} & 1       & $70.6\pm0.6$          & $73.5\pm0.5^*$          & $\mathbf{78.6\pm1.3}^*$      \\
                                                                         & 10      & $84.6\pm0.3$          & $\mathbf{86.8\pm0.1}$  & $\mathbf{87.0\pm0.5}$       \\
                                                                         & 50      & $\mathbf{88.7\pm0.2}$ & $88.0\pm0.1^*$          & $\mathbf{88.8\pm0.4}$       \\ 
\midrule
\multirow{3}{*}{SVHN}                                                    & 1       & $31.2\pm1.4$          & $\mathbf{57.3\pm0.1}^*$ & $52.2\pm2.2^*$               \\
                                                                         & 10      & $\mathbf{79.2\pm0.5}$ & $75.0\pm0.1$           & $74.9\pm0.4$                \\
                                                                         & 50      & $\mathbf{84.4\pm0.4}$ & $80.5\pm0.1$           & $80.9\pm0.3^*$               \\ 
\midrule
\multirow{3}{*}{CIFAR-10}                                                & 1       & $28.8\pm0.7$          & $49.9\pm0.2$           & $\mathbf{53.6\pm1.2}^*$      \\
                                                                         & 10      & $52.1\pm0.5$          & $62.7\pm0.3$           & $\mathbf{66.3\pm0.5}^*$      \\
                                                                         & 50      & $60.6\pm0.5$          & $68.6\pm0.2$           & $\mathbf{71.1\pm0.4}$       \\ 
\midrule
\multirow{2}{*}{CIFAR-100}                                               & 1       & $13.9\pm0.3$          & $15.7\pm0.2$           & $\mathbf{26.3\pm1.1}^*$      \\
                                                                         & 10      & $32.3\pm0.3$          & $28.3\pm0.1$           & $\mathbf{33.0\pm0.3}^*$      \\
\bottomrule
\end{tabular}
}
\label{tab:finite_network_transfer}
% \vspace{300mm}
% \vglue 1in
% \vskip{30mm}
\end{table}

\textbf{Network Structure and Training Setup.} Similar to previous work on dataset distillation, we used standard ConvNet architectures with three convolutional layers with average pooling and ReLU activations \citep{zhao2021DC,zhao2020differentiable,KIP2}. Similar to KIP \citep{KIP2}, we do not use instancenorm layers because of the lack of an infinite-width analog. During training, we used $N = 8$ random models, each with $C = 256$ convolutional channels per layer, and during test-time, we evaluated the datasets using the exact NNGP kernel using the neural-tangents library \citep{neural_tangents}. We consider both the fixed and learned label configurations, with Platt scaling applied and no data augmentation. We used the regularized Zero Component Analysis (ZCA) preprocessing for SVHN, CIFAR-10, and CIFAR-100, to improve KRR performance for color image datasets \citep{zca_good, KIP2}. More details are available in \cref{app:implementation}.

\textbf{Baselines.} We compare RFAD to recently developed advanced dataset distillation algorithms such as: KIP \citep{KIP1,KIP2}, Dataset Condensation with gradient matching (DC) \citep{zhao2021DC}, and differentiable  Siamese  augmentation (DSA) \citep{zhao2021dsa}.

Table \ref{tab:accuracy_plot} summarizes the results. We observe that in the fixed label configuration, our method outperforms other models in almost every dataset. In particular, it outperforms KIP by up to $6.1\%$ in the CIFAR-10 10 img/cls setting. We attribute this gain primarily to the use of Platt scaling. RFAD falls slightly behind KIP with learned labels. While this could partially be explained because we did not apply data augmentation, which marginally elevated performance for KIP on some datasets \citep{KIP2}, we hypothesize that the performance difference is caused by the increased gradient variance associated with the random feature method. Nevertheless, in all experiments, RFAD is at least two orders of magnitude faster than KIP (Figure \ref{fig:time_per_epoch}).

\subsection{Time Savings during training} 
\begin{wrapfigure}[15]{r}{0.5\textwidth}
\vspace{-7mm}
\begin{center}
\includegraphics[width=0.5\columnwidth]{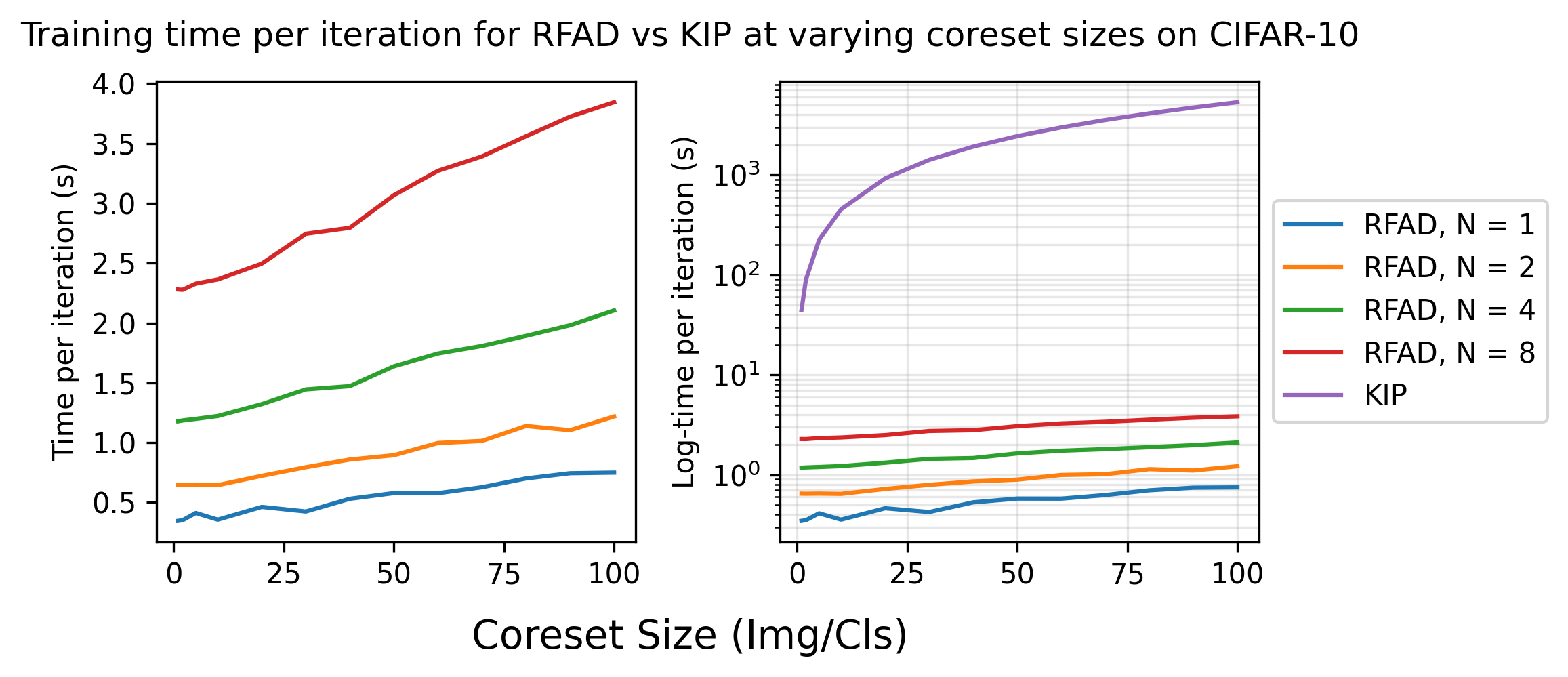}
\vspace{-6mm}
\caption{Time per training iteration for RFAD and KIP with varying number of models, $N$. Left: Linear plot of time. Right: Logarithmic time for training iteration. RFAD achieves over two-orders-of-magnitude speedup compared KIP per training iteration while converging with a similar number of iterations.}
\label{fig:time_per_epoch}
\end{center}
\vskip -0.2in
\end{wrapfigure}

Next, we evaluated the time efficiency of RFAD. \cref{fig:time_per_epoch} shows the time taken per training iteration on CIFAR-10 over coreset sizes and the number of models, $N$ used to evaluate the empirical NNGP kernel during training. Each training iteration contains 5120 examples from the training set.
\cref{fig:time_per_epoch} depicts that the time taken by RFAD is linear in both the number of models used during training and in the coreset size, validating the time complexity described above. We expect that for larger coreset sizes, the matrix inversion will begin to dominate due to its cubic complexity, but for small coreset sizes, the computation of the kernel matrix dominates the computation time. 

In the right-hand side plot in \cref{fig:time_per_epoch} we show the same plot in log-scale, compared to KIP. For KIP, we used a batch size of 5000, and rather than measuring the time taken, we use the calculation provided in appendix B of \citep{KIP2}, which describes the running time of the algorithm. We observe evidently that even for the modest coreset sizes, the quadratic time complexity of computing the exact kernel matrix in KIP results in it being multiple orders of magnitude slower than our RFAD. Both KIP and RFAD converge in between 3000-15000 training iterations, resulting in times between 1-14hrs for RFAD and several hundred GPU hours for KIP, depending on the coreset size dataset, and when the early stopping condition is triggered.

\subsection{NTK Kernel and Finite Network Transfer}

\begin{figure}[t]
\centering
\includegraphics[width=\textwidth]{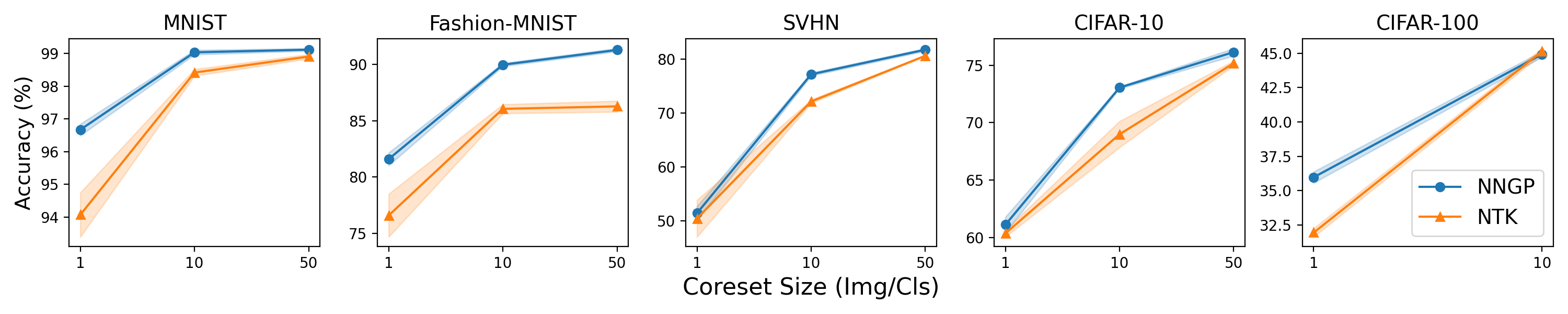}
\caption{NNGP to NTK transfer performance on RFAD distilled images. The blue line indicates the performance of RFAD distilled images evaluated on NNGP. The orange line shows the same images evaluated using NTK. Despite being trained using the empirical NNGP kernel, these images still perform well on the NTK kernel, losing at most a few percentage points. 
(n = 4)}
\label{fig:nngp_to_ntk}
\end{figure}

One of the key elements of the RFAD algorithm is the replacement of NTK with the empirical NNGP kernel. While we argued earlier that the two should exhibit a similar performance given their similar formalism, in this section, we verify this claim experimentally. We evaluated our distilled coresets obtained from our RFAD algorithm in two different transfer scenarios. In the first setting, at test time, we used an NTK kernel instead of the NNGP kernel. In the second setting, we trained a finite-width network with gradient descent on the distilled datasets obtained via RFAD. Similar to \citep{KIP2}, we used a 1024-width finite network for our finite-transfer results since it better mimics the infinite width setting that corresponds to the NTK.

Remarkably, as shown in \cref{fig:nngp_to_ntk}, in most datasets, these coresets suffer little to no performance drop when evaluated using NTK compared to the exact NNGP kernel, despite being trained using the empirical NNGP kernel. The largest performance gap is 8\% on SVHN with 10 images per class, and in some datasets, notably CIFAR-100, 10 img/cls evaluating using the NTK kernel outperforms NNGP. This suggests that either the exact NNGP kernel or the random feature NNGP kernel could potentially be used as a cheaper approximation for the exact NTK kernel. 

\cref{tab:finite_network_transfer} shows the resulting finite network transfer when training with gradient descent on our coresets. Our images appear to have the best performance in finite-network transfer, outperforming KIP in almost all benchmarks and the DC/DSA algorithms in many, despite DC/DCA being designed specifically for finite-width networks. We attribute this performance gain over KIP primarily to two tricks we used during training. Firstly, we applied centering, which, rather than training a typical network $f_{\theta}(x)$, we instead train a network with its output at initialization subtracted: $f_{\theta}(x) - f_{\theta_0}(x)$.

\begin{figure}[t]
\begin{center}
\centerline{\includegraphics[width=0.8\columnwidth]{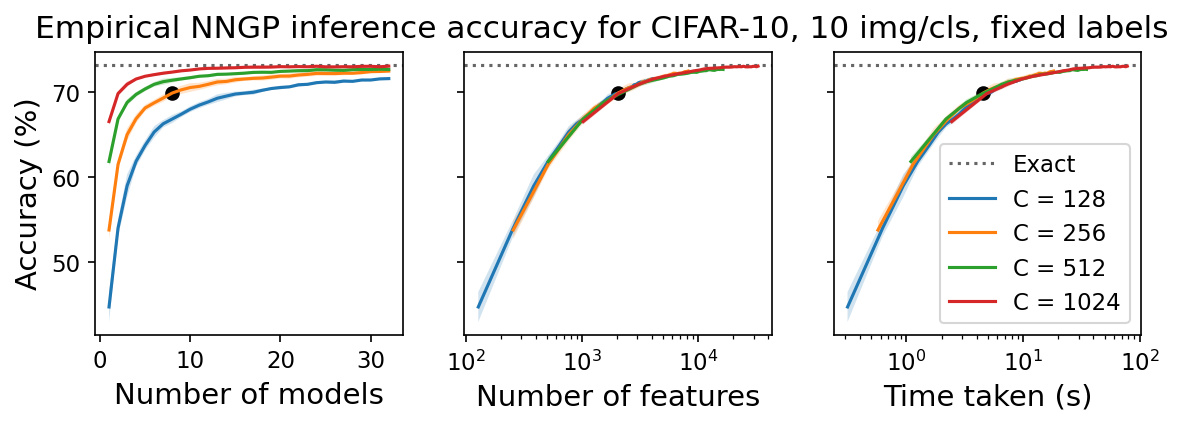}}
\caption{Empirical NNGP kernel performance at test-time with a varying number of models used to compute the empirical NNGP kernel and number of convolutional channels per model. %Increasing the model count improves performance, and the performance is dependent almost entirely on the total number of features used. The performance of finite networks closely matches exact computation for large feature counts. 
(n = 5)}
\label{fig:finite_network_inference}
\end{center}
\end{figure}

This has been shown empirically to speed up the convergence of finite-width networks by reducing the bias caused by the finite-width initialization while still preserving the NTK \citep{finite_vs_infinite, funny_l2}. We find that for these small datasets, this modification significantly improves performance. The second trick is label scaling; we scale the target labels by a factor $\alpha > 1$: $\mathcal{L}_{\alpha} = ||f_{\theta}(x) - \alpha y||_2^2/\alpha^2$, and at inference divide the model's outputs by $\alpha$. Note that this does not affect the infinite-width setting, as in KRR, the output is linear w.r.t. the support set labels. Ablations of these changes are in \cref{app:finite_network_ablations}. 

\begin{figure}[t]
\begin{center}
\centerline{\includegraphics[width=0.7\columnwidth]{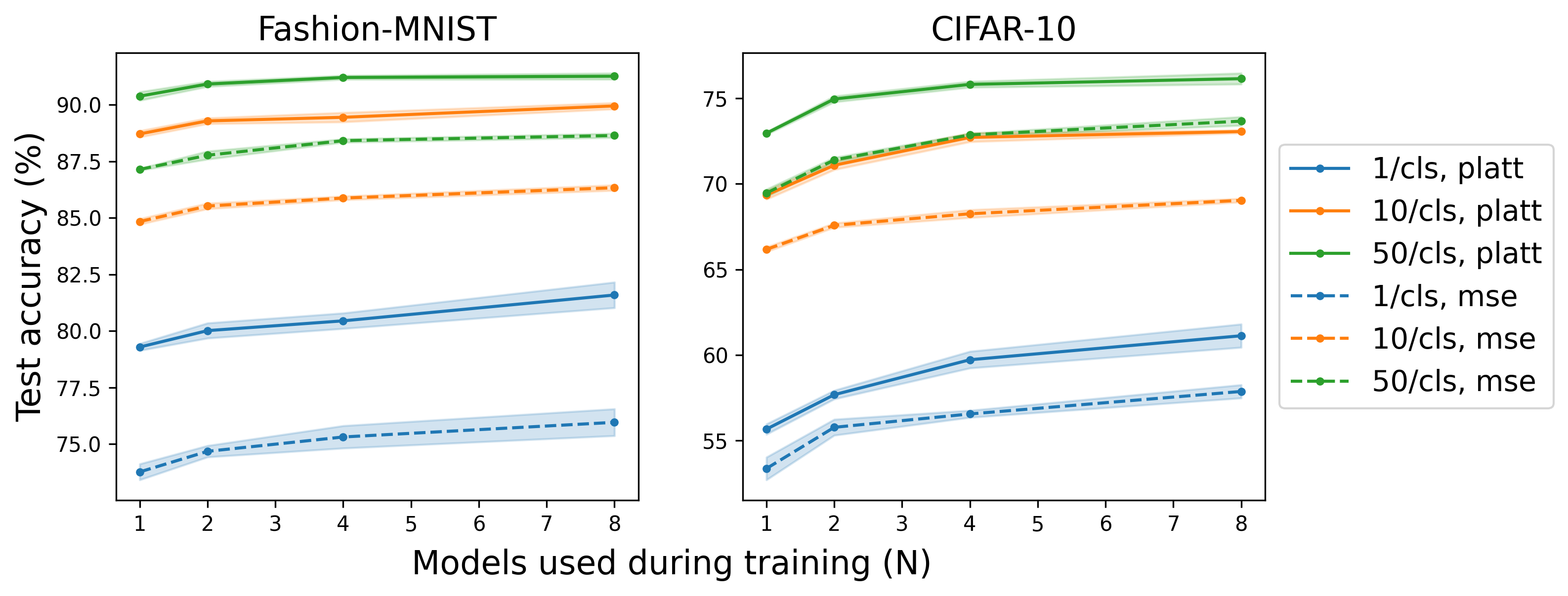}}
\caption{Objective function sensitivity. (n = 4) to use Platt scaling and the number of models during training (N) for Fashion-MNIST and CIFAR-10}
\label{fig:hypers_during_training}
\end{center}
\vskip -0.2in
\end{figure}

\subsection{Empirical NNGP Performance at Inference}
\label{sec:finite_at_inference}
To validate the efficacy of our method, we evaluated our distilled coresets using features from random networks as opposed to the exact kernel. We varied the width of individual networks between 128 and 1024 channels and the number of models between 1 and 32. \cref{fig:finite_network_inference} shows the resulting classification accuracy on the CIFAR-10 dataset with 10 images/class. The black dot represents the configuration we used during training: 8 models, each with width 256 (More ablations are provided in \cref{app:sec_empirical_inference}). We conclude that the random feature method, for all network widths, is able to reach close to the exact NNGP kernel performance (dotted line) if a sufficient number of models are used.
Interestingly, the performance is almost entirely dependent on the total number of features (proportional to $C \times N$, with $C$ being the number of convolutional channels) and not the width of individual networks, suggesting that the finite-width bias associated with random finite networks is minimal. 

In \cref{app:sec_empirical_inference}, we show that this can be taken to the extreme, with $70\%$ accuracy achieved with a network with a \textit{single} convolutional channel. These results corroborate the findings of \citep{GP_convolution1}, which first proposed this random feature method, where they found, like us, that performance was almost entirely determined by the total feature count.

Platt scaling. We performed ablations on the use of the cross-entropy loss and the number of models used during training. We reran our algorithm on CIFAR-10 and Fashion-MNIST, using either 1, 2, 4, or 8 models during training, using MSE loss or cross-entropy loss. \cref{fig:hypers_during_training} shows the resulting performance of these configurations. Evidently, using a cross-entropy loss results in substantial performance gains, even as much as $8\%$ as with Fashion-MNIST with one img/cls. %In contrast, our algorithm appears to be very robust to the number of models used during training, particularly for Fashion-MNIST. As shown in \cref{fig:time_per_epoch}, the time per epoch is proportional to $N$, meaning that by using 2 models as opposed to 8 during training, we could expect a 4x speedup with a minimal performance drop.

\section{RFAD Application I: Interpretability}
\label{sec:interpretability}

Large datasets contribute to the difficulty of understanding deep learning models. In this paper, we consider interpretability in the sense of the influence of individual training examples on network predictions \citep{hasani2019response,lechner2020neural,wang2022interpreting}. One method of understanding this effect is the use of influence functions, which seek to answer the following counterfactual question: which item in the training set, if left out, would change the model's prediction the most \citep{old_influence, understanding_black_bow_with_influence, influential_neighbors}? For deep networks, this can only be answered approximately. This is because retraining a network on copies of the training set with individual items left out is computationally intractable. One solution is to use kernel ridge regression on a small support set. We can recompute the KRR on the kernel matrices with the $i$th individual coreset element removed, with $K_{x, S\setminus i}$ $K_{S\setminus i, S \setminus i}$ being the resulting kernel matrices with the $i$th row/column corresponding to the $i$th coreset entry removed.

\begin{figure}[t]
\centering
\includegraphics[width=1\textwidth]{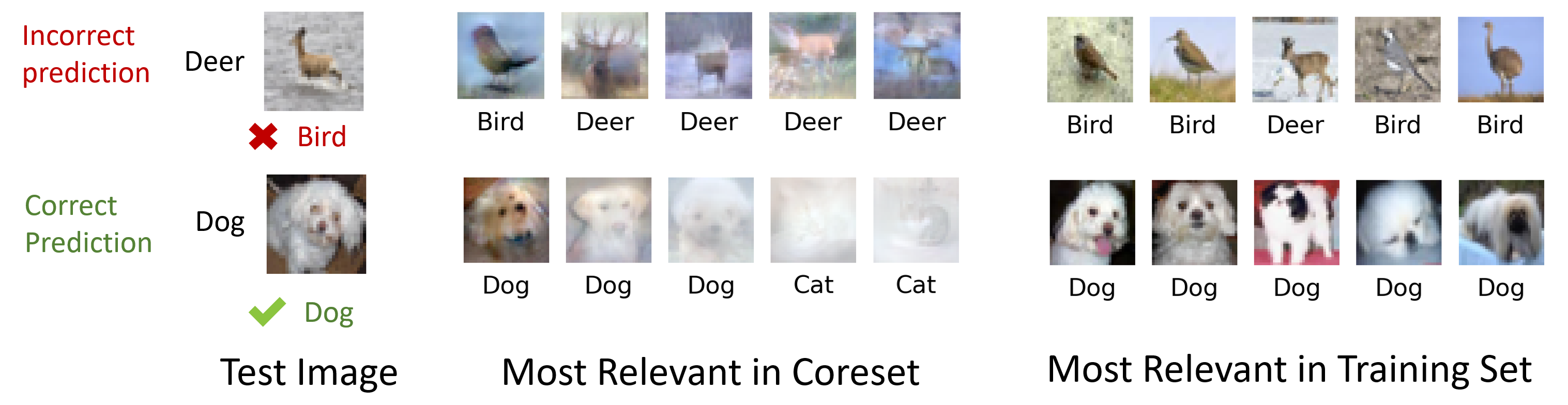}
% \vspace{-2mm}
\caption{The most relevant images for an incorrect (top row) and correct (bottom row) prediction on CIFAR-10. The most relevant coreset images are picked based on the coreset influence score $I$, and for their training set, the training influence score $J$. These metrics are fast to compute and result in semantically meaningful explanations for these two predictions.}
\label{fig:interp_demo}
\end{figure}

In particular, let $p(y_{\tiny{\textrm{test}}} = c| S)$ be the probability prediction (computed by applying Platt scaling) of an example belonging to class $c$ computed on the entire coreset, $S$. Let $p(y_{\tiny{\textrm{test}}} = c | S\setminus i)$ be the same prediction calculated with the $i$th coreset element removed. We define the influence score, $I_{i}$ of coreset element $i$ on $x_{\tiny{\textrm{test}}}$ as $\sum_{c \leq C} |p(y_{\tiny{\textrm{test}}} = c| S) - p(y_{\tiny{\textrm{test}}} = c | S\setminus i)|$. 
Taking the top values of $I_i$ yields the most relevant examples.

While this method provides a simple way of gaining insights into how a prediction depends on the coreset, it does not provide insight into how this prediction comes from the original training set which produced the coreset. The method can be extended to accommodate this. Heuristically, we conjecture that two elements are similar if their predictions depend on the same elements in the coreset. We compute $p(y_{j} = c| S)$ and $p(y_{j} = c| S\setminus i)$ for every element $j$ in the training set and $i$ in the coreset. Then, we define its influence embedding as $z^j_{i,c} = p(y_{j} = c| S) - p(y_{j} = c| S\setminus i), z^j \in R^{|S| \times |C|}$. This way, $z^j$ defines the sensitivity of a training datapoint prediction on the coreset. We compute the same embedding for a test datapoint $z^{\tiny{\textrm{test}}}$, and to compare data points we use cosine similarity, $J_{\tiny{\textrm{test}}, j} = \cos(z^{\tiny{\textrm{test}}}, z^j)$. Values of $z^j$ can be precomputed for the training set, typically in a few minutes for CIFAR-10, allowing for relatively fast queries, in contrast to the more expensive Hessian-inverse vector product required in \citet{understanding_black_bow_with_influence}, which is costly to compute and difficult to store.

\begin{wrapfigure}[19]{r}{0.5\textwidth}
\centering
\vspace{-4mm}
\includegraphics[width=0.5\columnwidth]{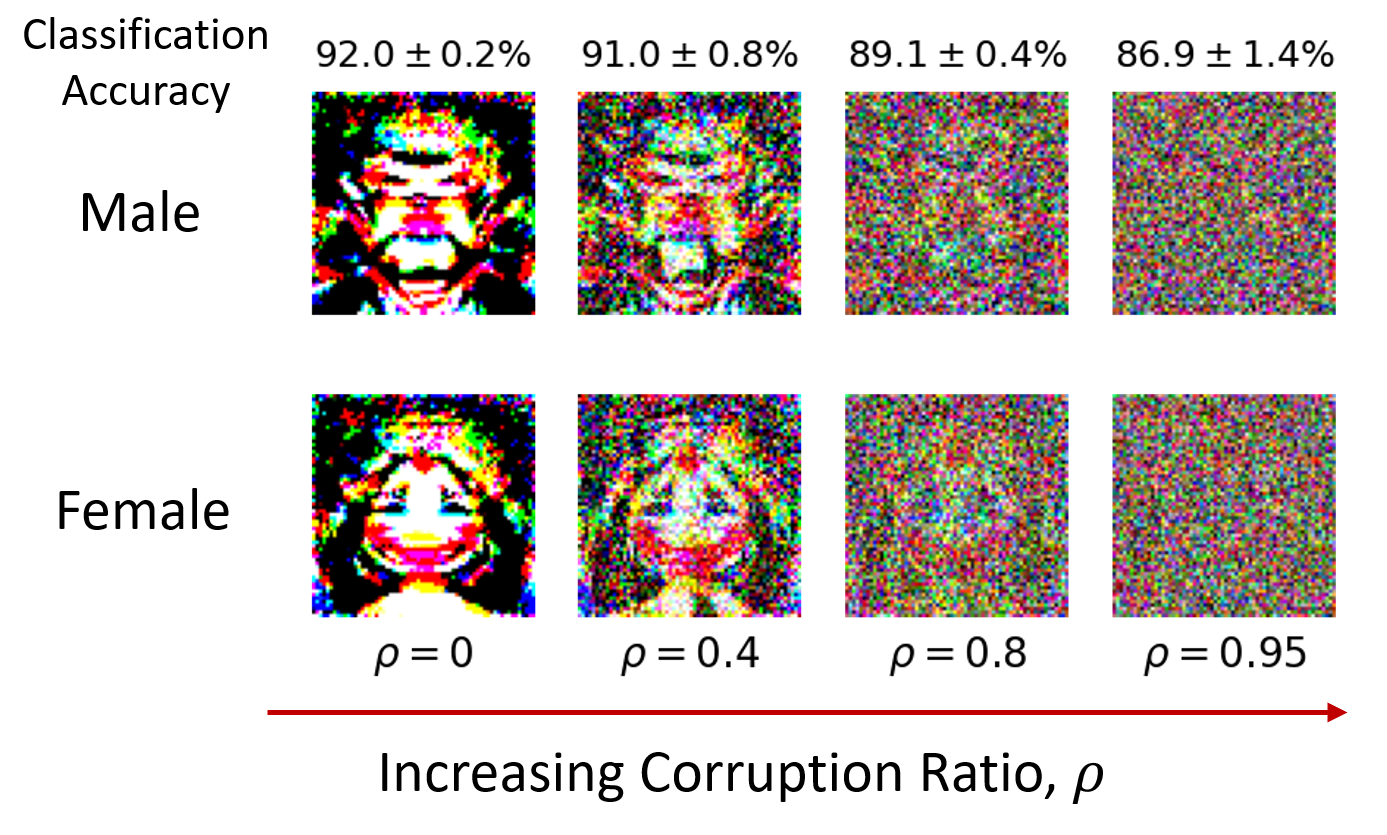}
\caption{CelebA distilled datasets for male/female classification with 1 image per class at varying corruption ratios. At $\rho = 0$, the distilled images are very interpretable, but at $\rho = 0.95$, the images look like white noise, despite achieving 86.9\% accuracy on the classification task.}
\label{fig:celeb_corruption_timeline}
\end{wrapfigure}
\cref{fig:interp_demo} shows the results of this algorithm applied to CIFAR-10 with 50 img/cls for an incorrectly and correctly predicted image. In both cases, the resulting queries are visually similar to the test data point. One could use this information to not only explain a single incorrect prediction but to understand harmful items in their test set, or where more data needs to be collected.

As a second application of RFAD, we create coresets that contain no human-understandable information from the source dataset yet still retain high test performance. \citep{KIP1} proposed the concept of a $\rho$-corrupted coreset: A fraction $\rho$ of the coresets elements are completely independent of the source dataset. Practically, for our algorithm, this means initializing the coreset with random noise and keeping a random $\rho$ fraction of the pixels kept at their initialization. We term this algorithm RFAD$_\rho$. Adding noise to gradient updates of the inputs of a network can be shown to give differentially private guarantees \citep{differential_privacy}. 
While our scheme does not provide the same guarantees, we note the following two privacy-preserving properties of RFAD$_\rho$: firstly, the distillation process is irreversible: there are many datasets for which a distilled dataset provides zero loss. %The distilled dataset achieves zero loss on itself, meaning that the original full dataset cannot be restored from a distilled set. 
Secondly, if the true data distribution assigns a low probability to images of white noise, then for high values of $\rho$, this guarantees that the distilled dataset stays far away in $L_2$ norm from real data points, since $\rho$ fraction of the pixels are stuck at their initialization. This means that a distilled RFAD$_\rho$ dataset will not recreate any real points in the training set.

\section{RFAD Application II: Privacy}
\label{sec:privacy}

\begin{wrapfigure}[16]{r}{0.5\textwidth}
\vspace{-7mm}
\begin{center}
\centerline{\includegraphics[width=0.5\columnwidth]{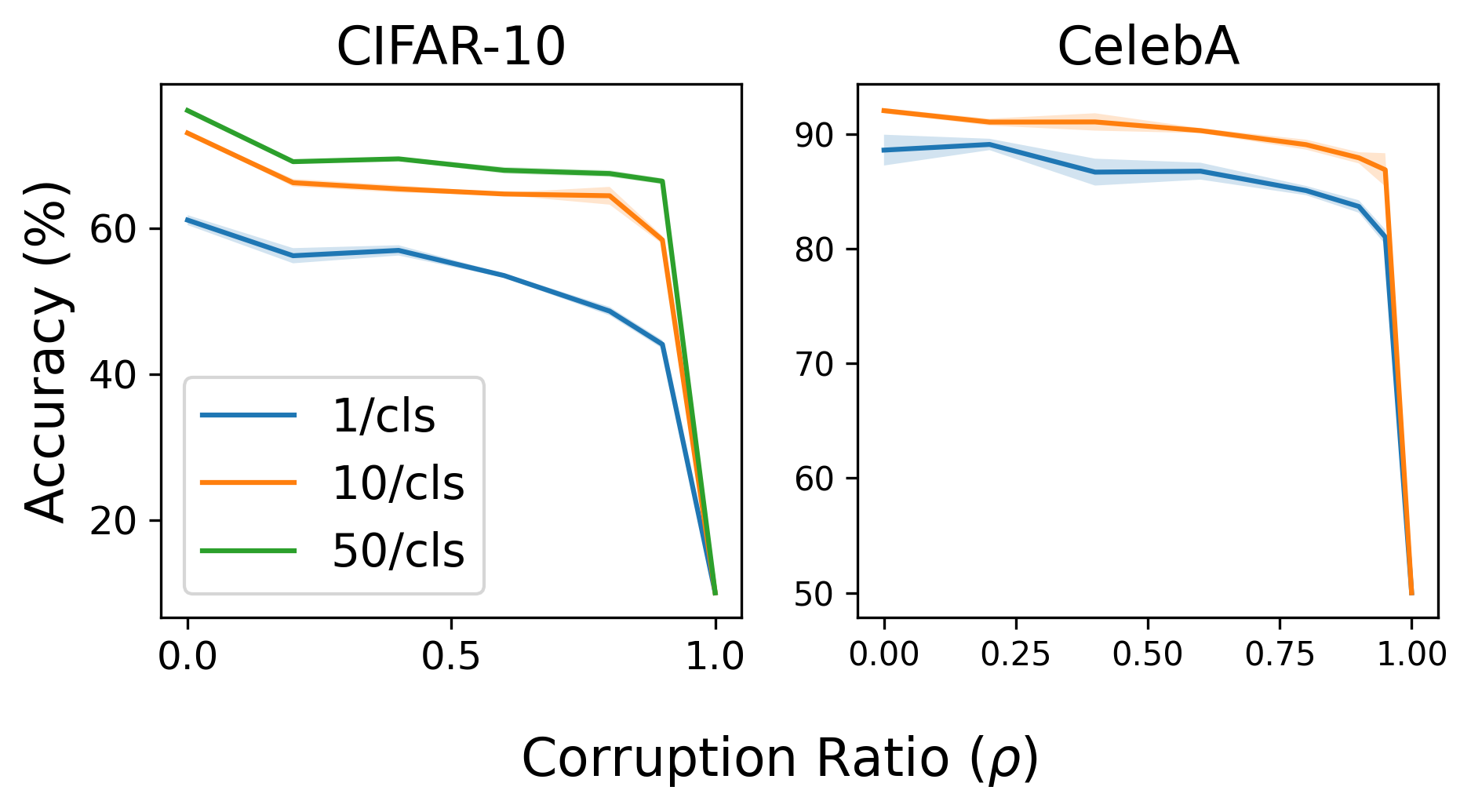}}
\caption{Performance of RFAD on CIFAR-10 and CelebA classification with varying support set sizes and corruption ratios. Performance degrades very gradually as noise is increased, still achieving high performance with 90\% corruption. (n = 3)}
\label{fig:corruption_combined}
\end{center}
\vskip -0.2in
\end{wrapfigure}
We applied RFAD$_\rho$ on CIFAR-10, and CelebA faces datasets. For CIFAR-10 we distilled the standard 10-class classification task, with corruption ratios taking values of $[0, 0.2, 0.4, 0.8, 0.9]$, with 1, 10 or 50 images per class. For CelebA, we performed male/female binary classification with corruption ratios between 0 and 0.95 with 1, 10, or 50 samples per class. \cref{fig:corruption_combined} show the resulting performance.

For CIFAR-10, even at corruption ratios of 0.9, we are able to achieve 40.6\% accuracy with one sample per class, far above the natural baseline of 16.1\% (\citep{KIP2} table A1). For CelebA, we achieve 81\% accuracy with only two images, one male and one female, with 95\% of the pixels in the image being random noise. We additionally visualize the distilled images for the male and female classes in the CelebA distillation task with one image per class in \cref{fig:celeb_corruption_timeline} at varying corruption ratios. While the image initially contains visually interpretable features with $\rho = 0$, they quickly devolve into pure noise at $\rho = 0.95$.

\section{Conclusions}

We proposed RFAD, a dataset distillation algorithm that provides a 100-fold speedup over the current state-of-the-art KIP while retaining accuracy. The speedup is primarily due to the use of the approximate NNGP kernel as opposed to the exact NTK one, reducing the time complexity from $O(|S|^2)$ to $O(|S|)$. The success of the approximation here, combined with the similarity between the NTK and NNGP kernels, suggests the random network NNGP approximation as an efficient method for algorithms where the exact computation of the NNGP or NTK kernel is infeasible. We analyzed our method comprehensively and showed its effectiveness, and proposed two applications in model interpretability and privacy preservation. With this new tool, we hope that future work could begin to use Neural Tangent Kernel as an algorithmic design tool in addition to its current theoretical use for neural network analysis. Lastly, RFAD has the following limitations:

\textbf{Use of instancenorm.}
In practice, we found that our datasets distilled without instancenorm do not transfer well to finite networks with it. Conversely, if we use random networks with instancenorm in RFAD, these transfer to finite networks with instancenorm but not to ones without or the NNGP kernel. This suggests that the features used by networks with/without instancenorm differ, making it difficult to distill datasets that perform well on both. We discuss this further in \cref{app:sec_instancenorm}.

\textbf{Overfitting in dataset distillation.}
On Platt scaling, we argued that the heavily constrained nature of dataset distillation leads to underfitting of the training set when using an MSE-style loss in KIP, and we verified the efficacy of using a Platt loss instead. However, we observed that in simple datasets, such as MNIST, or with large coresets relative to the data, such as CIFAR-100 with 10 images per class, we could overfit to the dataset. We found that these distilled datasets were able to achieve near 100\% classification accuracy on the training set, meaning that it was distilled perfectly in terms of the Platt-loss. This implies that adding more images would not improve performance. Thus we hypothesize that using a Platt loss would be detrimental if the compression ratio is low.

\section*{Acknowledgements}
%This research has been filed as a U.S. Provisional Patent, with Application No. 63/390,952 on July 20, 2022. All rights reserved by the authors.
This research has been funded in part by the Office of Naval Research Grant Number Grant N00014-18-1-2830 and the J. P. Morgan AI Research program. We are very grateful.
%}

%\bibliographystyle{plain}
%\bibliography{references}

\section*{Checklist}

\begin{enumerate}

\item For all authors...
\begin{enumerate}
  \item Do the main claims made in the abstract and introduction accurately reflect the paper's contributions and scope?
    \answerYes{}
  \item Did you describe the limitations of your work?
    \answerYes{}
  \item Did you discuss any potential negative societal impacts of your work?
    \answerNA{}
  \item Have you read the ethics review guidelines and ensured that your paper conforms to them?
    \answerYes{}
\end{enumerate}

\item If you are including theoretical results...
\begin{enumerate}
  \item Did you state the full set of assumptions of all theoretical results?
    \answerNA{}
        \item Did you include complete proofs of all theoretical results?
    \answerNA{}
\end{enumerate}

\item If you ran experiments...
\begin{enumerate}
  \item Did you include the code, data, and instructions needed to reproduce the main experimental results (either in the supplemental material or as a URL)?
    \answerYes{}
  \item Did you specify all the training details (e.g., data splits, hyperparameters, how they were chosen)?
    \answerYes{}
        \item Did you report error bars (e.g., with respect to the random seed after running experiments multiple times)?
    \answerYes{}
        \item Did you include the total amount of compute and the type of resources used (e.g., type of GPUs, internal cluster, or cloud provider)?
    \answerYes{}
\end{enumerate}

\item If you are using existing assets (e.g., code, data, models) or curating/releasing new assets...
\begin{enumerate}
  \item If your work uses existing assets, did you cite the creators?
    \answerYes{}
  \item Did you mention the license of the assets?
    \answerNA{}
  \item Did you include any new assets either in the supplemental material or as a URL?
    \answerNA{}
  \item Did you discuss whether and how consent was obtained from people whose data you're using/curating?
    \answerNA{}
  \item Did you discuss whether the data you are using/curating contains personally identifiable information or offensive content?
    \answerNA{}
\end{enumerate}

\item If you used crowdsourcing or conducted research with human subjects...
\begin{enumerate}
  \item Did you include the full text of instructions given to participants and screenshots, if applicable?
    \answerNA{}
  \item Did you describe any potential participant risks, with links to Institutional Review Board (IRB) approvals, if applicable?
    \answerNA{}
  \item Did you include the estimated hourly wage paid to participants and the total amount spent on participant compensation?
    \answerNA{}
\end{enumerate}

\end{enumerate}

%%%%%%%%%%%%%%%%%%%%%%%%%%%%%%%%%%%%%%%%%%%%%%%%%%%%%%%%%%%%

\newpage
\appendix

\section*{Appendix}

\section{Code}
The code is available at \texttt{https://github.com/yolky/RFAD}

\section{Notations}
\begin{table}[ht]
\centering
\begin{tabular}{cc}\toprule
Symbol & Meaning                    \\\midrule
$H$  & Input image height in pixels\\
$W$       &                  Input image width in pixels    \\
$D$       &       Network Depth              \\
$C$       &       Number of convolutional channels (network width)\\
$N$       &       Number of models used during traning\\
$M$       &       Number of network features, proportional to $C$\\
$|T|$       &       Training set size\\
$|B|$       &       Training batch size\\
$|S|$       &       Support set/coreset size\\\bottomrule
\end{tabular}
\caption{Notation for variables used throughout the paper. In some cases, $C$ also refers to number of classes in the dataset (\cref{sec:interpretability}), however it is clear from context.}
\end{table}
\section{Time Complexity Discussion}
\label{app:more_time_complexity}

Computing the approximate kernel matrices for RFAD consists of three main steps: random feature calculation, random feature matrix multiplication, and kernel matrix inversion. KIP, in comparison consists of two main steps: kernel matrix computation, and kernel matrix inversion. In practice, random feature matrix multiplication and kernel matrix inversion require negligible amounts of time for our coreset sizes, so we instead focus on the slowest part for RFAD and KIP: random feature calculation and kernel matrix computation, respectively. We discuss the runtime per training iteration of our algorithm for RFAD and KIP, assuming training batch size $|B|$ and coreset size $|S|$. Additionally, we split the discussion into the $|B| \leq |S|$ and $|B| > |S|$ regimes, as the exact complexity differs between these settings.

\textbf{Case 1: RFAD with $|B| \leq |S|$:} $O(|S|)$ random feature computation dominates, while the  matrix computation (multiplying feature vectors together) is cheap. The resulting complexity is effectively $O(|S|)$.

\textbf{Case 2: KIP with $|B| < |S|$:} $O(|S|^2)$ computation of $K_{SS}$ is the dominant term, in which case the resulting time complexity is effectively $O(|S|^2)$.

\textbf{Case 3: RFAD with $|B| > |S|$:} $O(|S|)$ and $O(|B|)$  and  random feature computation for the coreset and training batch are both are relatively expensive. Again, multiplying the random feature matrices together is cheap. The resulting complexity is thus $O(\lambda_1|S| + \lambda_2|B|)$ Training batch random features can be computed more quickly than coreset random features as the computation graph does not need to be retained for backpropagation, so $\lambda_1 >> \lambda_2$.

\textbf{Case 4: KIP with $|B| > |S|$:} $O(|B||S|)$ computation of $K_{BS}$ is the dominant term, in which case the resulting time complexity is $O(|B||S|)$, meaning linear complexity in the coreset size. However, \textbf{since $|B| > |S|$, this is effectively worse than $O(|S|^2)$ scaling.} Therefore, KIP is only linear in a regime where linear scaling is worse than quadratic scaling, due to a high time coefficient associated with the linear term. Note that unlike RFAD, we cannot compute $K_{SS}$ entries any quicker than $K_{BS}$ entries, as we cannot factorize it into coreset and training batch features, meaning we must retain the entire computation graph.

\section{Implementation details}
\label{app:implementation}
\subsection{Preprocessing}
For black/white datasets, i.e., MNIST and Fashion-MNIST, we use standard preprocessing, where we subtract the mean and divide by the standard deviation.

For color dataset SVHN, CIFAR-10, CIFAR-100, we use regularized Zero Component Analysis (ZCA) preprocessing with a regularization parameter of $\lambda = 0.1$ for all datasets. A description of this preprocessing method is available in the appendix of \citep{KIP2}, or in \citet{zca_good}. For CIFAR-10 and CIFAR-100, we did not include the unit normalization step in regularized ZCA as we found it reduced performance by around 1-2\%.
\subsection{Architectures}
For all architectures, we use the same ConvNet architecture used in \citet{zhao2021DC, zhao2021dsa, KIP2} which consists of three layers of $3\times 3$ convolutions, $2\times 2$ average pool, and ReLU activations, followed by a fully connected layer. We do not use any normalization layers in any experiments unless otherwise stated.

For weight initialization, we use standard parameterization with Gaussian weight and bias initializations with variances $\sigma^2_w = 2$ and $\sigma^2_b = 0.1$, following \citep{KIP2}. Note that by default, PyTorch uses Kaiming uniform initializations, which means we had to write custom convolutional layers to have Gaussian initializations. While we did not collect data on this, we found this difference in initialization to be negligible - for all intents and purposes, the default initialization works just as well but corresponds to a slightly different NNGP process.

For RFAD training, we used neural networks with 256 convolutional channels per layer. Additionally, we removed the final fully-connected layer and used the representations after the final ReLU layer to calculate the NNGP kernel instead. This can be done by noting that for fully-connected layers $K^{l} = \sigma^2_w K^{l-1} + \sigma^2_b$. This theoretically removes some variance associated with the final layer and saves on some memory. In practice, we found this did not affect performance.
\subsection{Training}
\label{app:implementation_training}
We used Adabelief optimizer \citep{adabelief} with a learning rate of $1e-3$ for all experiments and parameters, and $\varepsilon = 1e-16$. 

Additionally, we found it useful to split up the representation of $X_S$ into two pieces: one parameter $\hat{X}_S$ with the same shape as $X_S$: $\mathbb{R}^{|S|\times C\times H\times W}$ and another matrix $T \in \mathbb{R}^{(C\times H\times W) \times (C\times H\times W)}$. For example, for CIFAR-10, $T$ would be a $3072\times 3072$ matrix ($3072 = 3 \times 32 \times 32$). Then, we compute $X_S$ as $X_S = \textrm{reshape}(T \textrm{flatten}(\hat{X}_S)$. Note this is like ZCA preprocessing where the transformation matrix is learned from $T$. In the code we refer to this as the \texttt{transform\_mat}.

Note that this does not add any extra variables for the coreset, but in practice, we found that this trick resulted in slightly faster convergence, particularly at initialization. The $T$ matrix is initialized at the identity and learned with a small learning rate (5e-5). While we do not have a theoretical justification for why this speeds up optimization, we believe it may allow the optimizer to learn dependencies between parameters, perhaps allowing it to behave more like a second-order optimizer.

The coreset is initialized with a class-balanced subset of the data. For the labels, we use one-hot vectors with $1/C$ subtracted so that the vector is zero-mean. E.g. for 10 samples per class the label would be $[0.9, -0.1, -0.1, ... -0.1]$.

In experiments that use Platt scaling, we learn the logarithm of $\tau$ with a learning rate of 1e-2.

For each gradient step/training iteration, we use 5120 training examples. We compute features for these in 4 batches of 1280. We save memory in this step by calculating these features with the \texttt{torch.no\_grad()} flag, as we do not backpropagate through these values. For the matrix inversion, we found it helpful to use double-precision since the process is sensitive to rounding errors.

Like with KIP, we used a adaptive kernel regularizer when computing the inverse in $(K_{SS} + \lambda I)^{-1}$. We parameterized this as $\lambda = \lambda_0 \bar{\textrm{tr}}(K_{SS})$, where $\bar{\textrm{tr}}$ is the average value along the diagonal. We used $\lambda_0 = 5e-3$.

We performed early stopping with the patience of 1000 iterations. Every 40 epochs, the loss on a validation set of size 1000 is measured. This validation set is a subset of the training set (we are training on the validation set); however, we used a fixed set of 16 random neural networks when calculating the validation loss.

For results in \cref{tab:accuracy_plot}, we ran the algorithm four separate times for each configuration.
\subsection{Runtime experiments}
\label{sec:app_time_taken}
The time taken was calculated by running RFAD on CIFAR-10 with images per class in $[1, 2, 5, 10, 20, 30, 40, 50, 60, 70, 80, 90, 100]$, averaging over 200 training iterations with $N = 1, 2, 4, 8$. These experiments are run on a single RTX 3090. To make the results comparable to a V100, we add a conservative 40\% extra time taken.
\subsection{Finite Network Transfer}
We use finite networks with 1024 convolutional channels with the same weight and bias-variance as we trained on, again with standard parameterization. We used SGD with a learning rate of either $1e-1, 1e-2, 1e-3, 1e-4$, momentum 0.9, weight decay of either $0, 1e-3$ and label scaling coefficients in $1, 2, 8, 16$. When weight decay was used, we used the modified weight decay $||\theta - \theta_0||^2_2$ given in \citet{funny_l2}, standard weight decay would not result in zero-output when centering is used. The best hyperparameters were determined by the best validation set accuracy (taking from the training set) on the first run of the algorithm. Unless otherwise stated, we use the centering trick. Batch sizes are up to 500, meaning that we performed full-batch gradient descent for all experiments except CIFAR-100 with 10 images per class.

The results reported in \cref{tab:finite_network_transfer} are the average of 12 training runs: 3 finite network training runs for each of the 4 repeat runs of RFAD.
\subsection{Privacy experiments}
For CIFAR-10 corruption experiments, we use the same training protocol as in \cref{app:implementation_training}. Rather than initializing the coreset with real images, we initialize with white noise in image space. The corruption constraint is applied in \textit{pixel} space rather than the ZCA space. Note that the corruption mask (whether a pixel can be optimized) is done per-pixel and per-channel, meaning that a pixel can have its red and green channels fixed but blue channel free.

We downsized the CelebA dataset to $64\times 64$ images, applying standard preprocessing for this experiment. We used a training batch size of 1280 for CelebA. The full dataset achieves an accuracy of 97.6\%.

For the privacy experiments, we ran only a single run for each corruption ratio. To calculate the means and standard deviations, we took the best iteration, based on the early stopping condition, and iterations at 400, 200 iterations before, and 200 and 400 after.

\section{Time taken additional results}
\label{app:sec_time_taken_extra}

As discussed in section 4.1 and \cref{sec:app_time_taken}, we added an extra 40\% to iteration times to make or times on a RTX 3090 comparable to a Tesla V100. In this section, we report the original times. Additionally, we report the total number of training iterations and total runtimes for all our distillation results. Note that because of the early stopping condition, the epochs used during evaluation are 1000 iterations before the iteration counts we report here.

\begin{figure}[ht]
\vskip 0.2in
\begin{center}
\centerline{\includegraphics[width=0.6\textwidth]{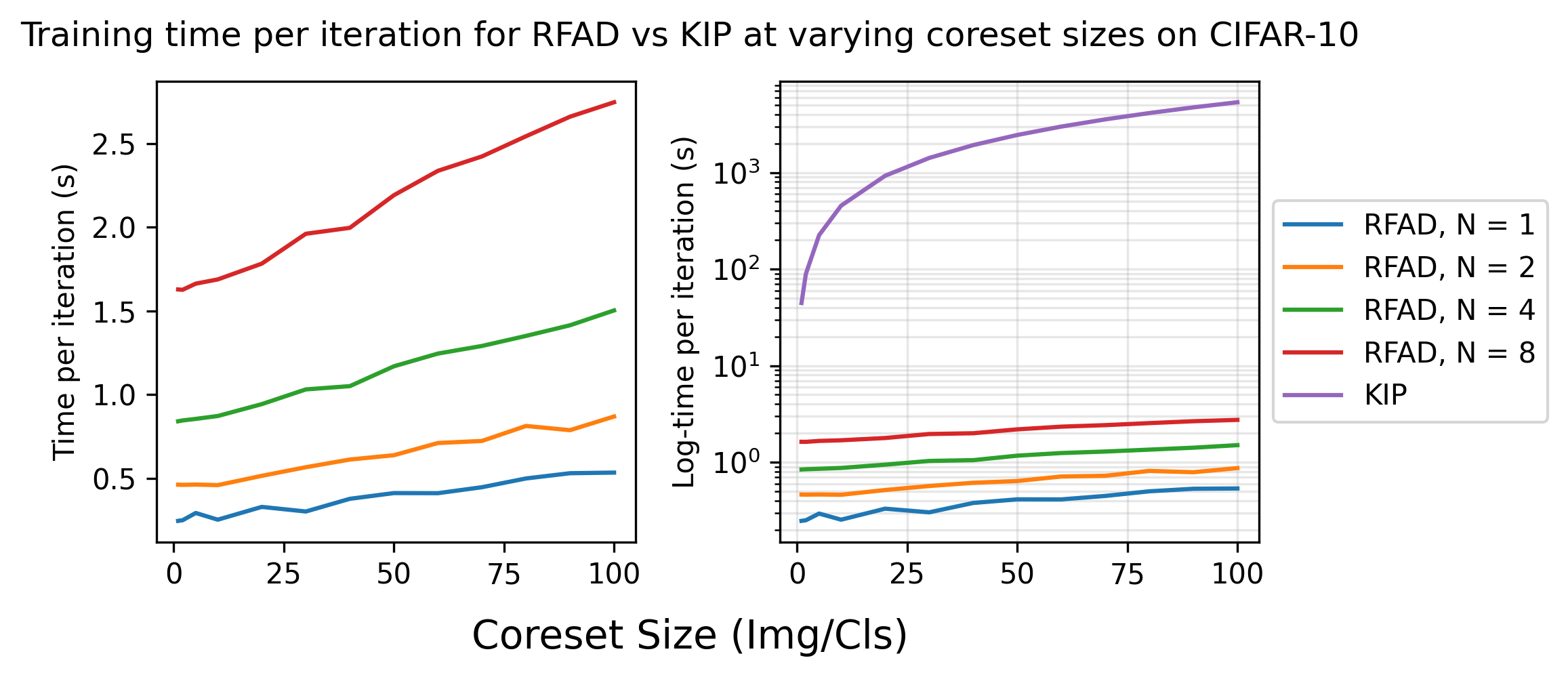}}
\caption{Time per training iteration for CIFAR-10 with varying number of models used during and and support sizes run on an RTX 3090 compared to KIP}
\end{center}
\vskip -0.2in
\end{figure}

\begin{table}
\centering
\caption{Total runtime and iteration for all RFAD distilled datasets. All experiments converge in under 10h on a single RTX 3090, with label learning usually taking longer.}
\resizebox{\textwidth}{!}{
\begin{tabular}{cccccccc} 
\toprule
                              &         & \multicolumn{3}{c}{Fixed Labels}                                               & \multicolumn{3}{c}{Learned Labels}                     \\
                              & Img/Cls & Number of iterations & \begin{tabular}[c]{@{}c@{}}Total elapsed\\time (h)\end{tabular} & \begin{tabular}[c]{@{}c@{}}Average time\\per iteration (s)\end{tabular} & Number of iterations & \begin{tabular}[c]{@{}c@{}}Total elapsed\\time (h)\end{tabular} &  \begin{tabular}[c]{@{}c@{}}Average time\\per iteration (s)\end{tabular}  \\ 
\midrule
\multirow{3}{*}{MNIST}         & 1       & $6650 \pm 859$ & $1.6 \pm 0.2$ & $0.86$    & $7390 \pm 1907$ & $1.7 \pm 0.5$ & $0.85$ \\
                              & 10      & $11070 \pm 3049$ & $2.8 \pm 0.8$ & $0.90$    & $12770 \pm 2609$ & $3.2 \pm 0.6$ & $0.90$ \\
                              & 50      & $8330 \pm 1225$ & $2.7 \pm 0.4$ & $1.18$    & $7420 \pm 1047$ & $2.4 \pm 0.3$ & $1.19$ \\
\midrule
\multirow{3}{*}{Fashion-MNIST} & 1       & $9580 \pm 1740$ & $2.3 \pm 0.4$ & $0.85$    & $9390 \pm 1924$ & $2.2 \pm 0.5$ & $0.85$ \\
                              & 10      & $14780 \pm 2025$ & $3.7 \pm 0.5$ & $0.89$    & $13010 \pm 2893$ & $3.2 \pm 0.7$ & $0.89$ \\
                              & 50      & $11190 \pm 2056$ & $3.6 \pm 0.7$ & $1.17$    & $12230 \pm 2206$ & $3.9 \pm 0.7$ & $1.15$ \\ 
 \midrule
\multirow{3}{*}{SVHN}          & 1       & $7540 \pm 1521$ & $3.4 \pm 0.7$ & $1.60$    & $5700 \pm 1267$ & $2.5 \pm 0.6$ & $1.59$ \\
                              & 10      & $9060 \pm 2155$ & $4.2 \pm 1.0$ & $1.67$    & $7330 \pm 1954$ & $3.4 \pm 0.9$ & $1.66$ \\
                              & 50      & $8270 \pm 2717$ & $4.8 \pm 1.5$ & $2.10$    & $10370 \pm 758$ & $6.0 \pm 0.5$ & $2.08$ \\
 \midrule
\multirow{3}{*}{CIFAR-10}      & 1       & $4610 \pm 778$ & $2.0 \pm 0.3$ & $1.56$    & $4300 \pm 1457$ & $1.9 \pm 0.6$ & $1.60$ \\
                              & 10      & $8310 \pm 2096$ & $3.8 \pm 1.0$ & $1.64$    & $9210 \pm 923$ & $4.3 \pm 0.4$ & $1.66$ \\
                              & 50      & $8370 \pm 2335$ & $4.9 \pm 1.3$ & $2.13$    & $14140 \pm 4901$ & $8.4 \pm 3.0$ & $2.13$ \\
 \midrule
\multirow{2}{*}{CIFAR-100}     & 1       & $9820 \pm 2067$ & $4.7 \pm 1.0$ & $1.71$    & $13650 \pm 4161$ & $6.1 \pm 2.3$ & $1.62$ \\
                              & 10      & $8410 \pm 2069$ & $6.3 \pm 1.5$ & $2.72$    & $13610 \pm 1463$ & $9.6 \pm 0.7$ & $2.54$ \\
\bottomrule
\end{tabular}
}
\end{table}

\section{Centering and label scaling for finite networks ablations}
\label{app:finite_network_ablations}

\cref{tab:app_finite_network_transfer_full} shows the full table finite network transfer results. We see that label scaling and centering provide performance benefits, particularly for small support sets. We do not have a theoretical explanation for why label scaling improves performance. \citep{lazy_training} suggests that label scaling for values $\alpha < 1$ should result in the network staying in the lazy regime, but in our case, we found values of $\alpha > 1$ to improve performance.

\begin{table}[ht]
% \onecolumn
\centering
\caption{Finite network transfer performance of KIP distilled images. We additional report performance if either label scaling or centering is not used. Label scaling and centering provide benefits particularly for smaller support set sizes.}
\begin{tabular}{cccccc} 
\toprule
                                                                         & Img/Cls & \begin{tabular}[c]{@{}c@{}}Fixed labels\\no label scale\end{tabular}            & \begin{tabular}[c]{@{}c@{}}Fixed labels\\no centering\end{tabular}           & Fixed labels               &  Learned labels\\
\midrule
\multirow{3}{*}{MNIST}                                                   & 1       & $84.1 \pm 5.5 $ & $90.1 \pm 3.3 $ & $92.2 \pm 2.1 $ & $94.4 \pm 1.5 $ \\
                                                                         & 10      & $96.7 \pm 0.3 $ & $98.3 \pm 0.2 $ & $98.4 \pm 0.2 $ & $98.5 \pm 0.1 $ \\
                                                                         & 50      & $98.5 \pm 0.2 $ & $98.8 \pm 0.1 $ & $98.8 \pm 0.1 $ & $98.8 \pm 0.1 $ \\ 
\midrule
\multirow{3}{*}{\begin{tabular}[c]{@{}c@{}}Fashion-\\MNIST\end{tabular}} & 1       & $51.0 \pm 19.2 $ & $69.8 \pm 3.4 $ & $76.7 \pm 1.7 $ & $78.6 \pm 1.3 $ \\
                                                                         & 10      & $84.5 \pm 0.9 $ & $85.0 \pm 0.7 $ & $87.0 \pm 0.5 $ & $85.9 \pm 0.7 $ \\
                                                                         & 50      & $87.7 \pm 0.4 $ & $87.4 \pm 0.4 $ & $88.8 \pm 0.4 $ & $88.5 \pm 0.4 $ \\
\midrule
\multirow{3}{*}{SVHN}                                                    & 1       & $32.6 \pm 3.0 $ & $40.2 \pm 2.9 $ & $43.1 \pm 2.4 $ & $52.2 \pm 2.2 $ \\
                                                                         & 10      & $65.7 \pm 1.0 $ & $72.9 \pm 0.8 $ & $73.6 \pm 1.0 $ & $74.9 \pm 0.4 $ \\
                                                                         & 50      & $78.0 \pm 0.3 $ & $78.5 \pm 0.3 $ & $80.1 \pm 0.4 $ & $80.9 \pm 0.3 $ \\
\midrule
\multirow{3}{*}{CIFAR-10}                                                & 1       & $48.7 \pm 1.6 $ & $48.0 \pm 1.7 $ & $53.2 \pm 1.2 $ & $53.6 \pm 0.9 $ \\
                                                                         & 10      & $63.4 \pm 0.6 $ & $56.9 \pm 1.0 $ & $66.1 \pm 0.5 $ & $66.3 \pm 0.5 $ \\
                                                                         & 50      & $69.5 \pm 0.4 $ & $68.0 \pm 0.6 $ & $71.1 \pm 0.4 $ & $70.3 \pm 0.5 $ \\
\midrule
\multirow{2}{*}{CIFAR-100}                                               & 1       & $19.6 \pm 0.6 $ & $21.2 \pm 0.3 $ & $24.2 \pm 0.4 $ & $26.3 \pm 1.1 $ \\
                                                                         & 10      & $30.5 \pm 0.3 $ & $18.5 \pm 0.6 $ & $30.3 \pm 0.3 $ & $33.0 \pm 0.3 $ \\
\bottomrule
\end{tabular}
\label{tab:app_finite_network_transfer_full}
\end{table}

\section{Effect of InstanceNorm}
\label{app:sec_instancenorm}

In \cref{sec:limitations} we discussed the observation that if we attempt to use instancenorm for finite network training for KIP distilled images, we do not see good performance. Conversely, if we use random networks with instancenorm in RFAD, these distilled datasets do not perform well on networks without instancenorm, either in the KRR case or finite network case. \cref{tab:app_instancenorm} shows the exact results. For NNGP KRR with instancenorm, we used the empirical NNGP kernel, with 32 networks with 1024 channels, as there is no exact implementation of the instancenorm NNGP in the neural-tangents PyTorch library \citep{neural_tangents}.

\begin{table}
\centering
\caption{Accuracy of RFAD distilled datasets run with or without networks with instancenorm during training evaluated on NNGP KRR and finite networks with SGD with or without instancenorm. We see that transferring from instancenorm to no instancenorm or vice versa incurs a large performance penalty. * indicates thats the emperical NNGP kernel was used, as there is no exact implementation of instancenorm in the neural-tangents library \citep{neural_tangents}}
\label{tab:app_instancenorm}
\begin{tabular}{cccccc} 
\toprule
                              &         & \multicolumn{2}{c}{Without InstanceNorm}                                               & \multicolumn{2}{c}{With InstanceNorm}                     \\
                              & Img/Cls & NNGP KRR & Finite Network & NNGP KRR* & Finite Network \\ 
\midrule
\multirow{3}{*}{\begin{tabular}[c]{@{}c@{}}Trained without\\InstanceNorm\end{tabular}}         & 1 &$61.1 \pm 0.7 $& $53.2 \pm 1.2 $& $35.3 \pm 0.9 $& $37.4 \pm 1.1 $\\
                              & 10      &$73.1 \pm 0.1 $& $66.1 \pm 0.5 $& $45.9 \pm 1.8 $& $45.1 \pm 1.1 $\\
                              & 50      &$76.1 \pm 0.3 $& $71.1 \pm 0.4 $& $59.1 \pm 0.4 $& $50.3 \pm 0.8 $\\
\midrule
\multirow{3}{*}{\begin{tabular}[c]{@{}c@{}}Trained with\\InstanceNorm\end{tabular}} & 1       &$18.1 \pm 3.7 $& $40.6 \pm 3.7 $& $57.8 \pm 0.7 $& $52.8 \pm 0.7 $\\
                              & 10      &$25.6 \pm 5.3 $& $36.3 \pm 1.5 $& $71.1 \pm 0.2 $& $63.5 \pm 0.5 $\\
                              & 50      &$52.5 \pm 0.5 $& $55.0 \pm 0.6 $& $74.4 \pm 0.2 $& $62.2 \pm 0.4 $\\
\bottomrule
\end{tabular}
\end{table}

\section{Interpretability additional examples}
\label{app:sec_interp_extra}

\begin{figure}[t]
\vskip 0.2in
\begin{center}
\centerline{\includegraphics[width=0.8\textwidth]{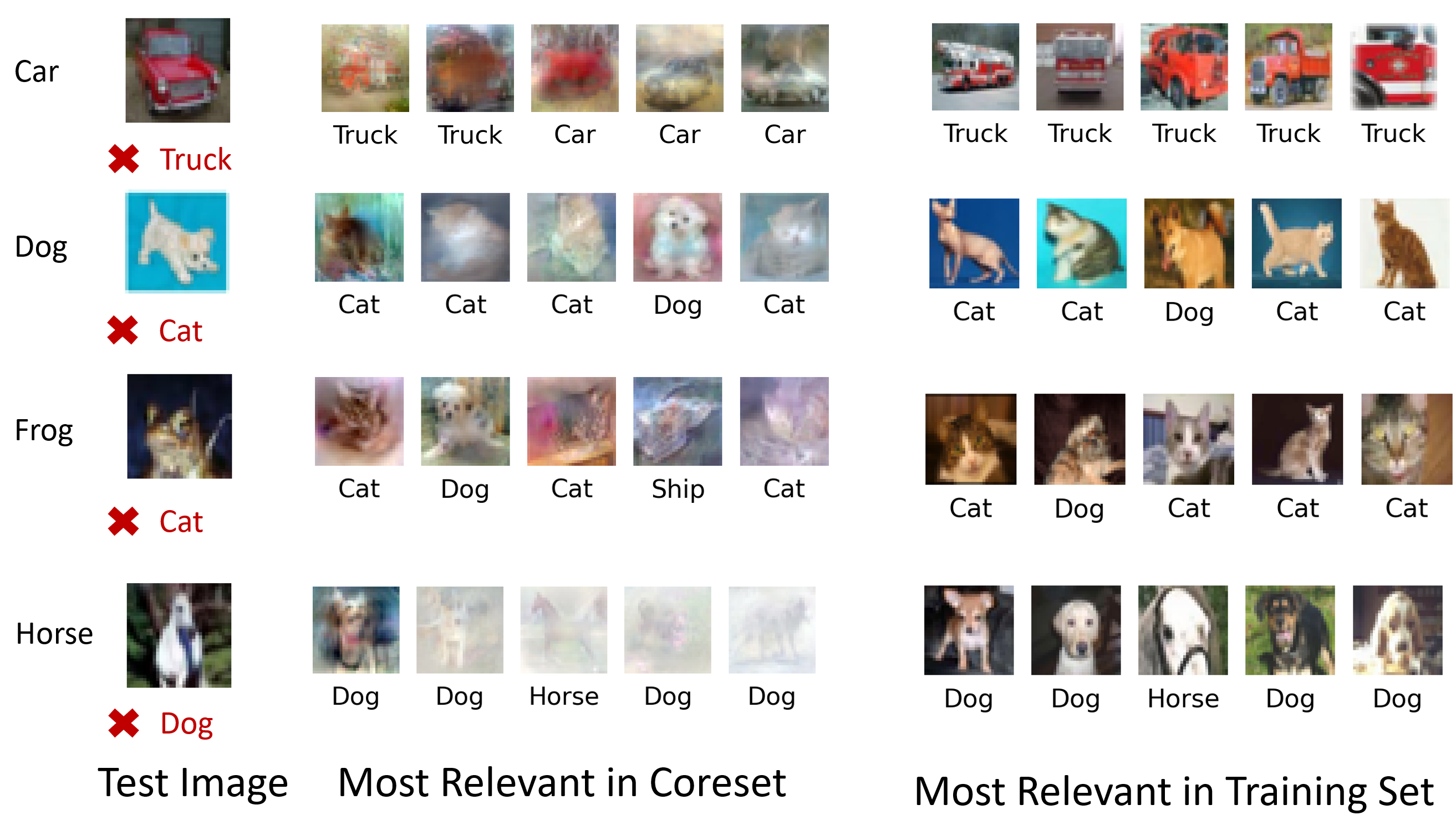}}
\caption{Incorrectly predicted test images and the most relevant images in the coreset and training set for CIFAR-10, 50 images/cls}
\end{center}
\vskip -0.2in
\end{figure}

\begin{figure}[ht]
\vskip 0.2in
\begin{center}
\centerline{\includegraphics[width=0.8\textwidth]{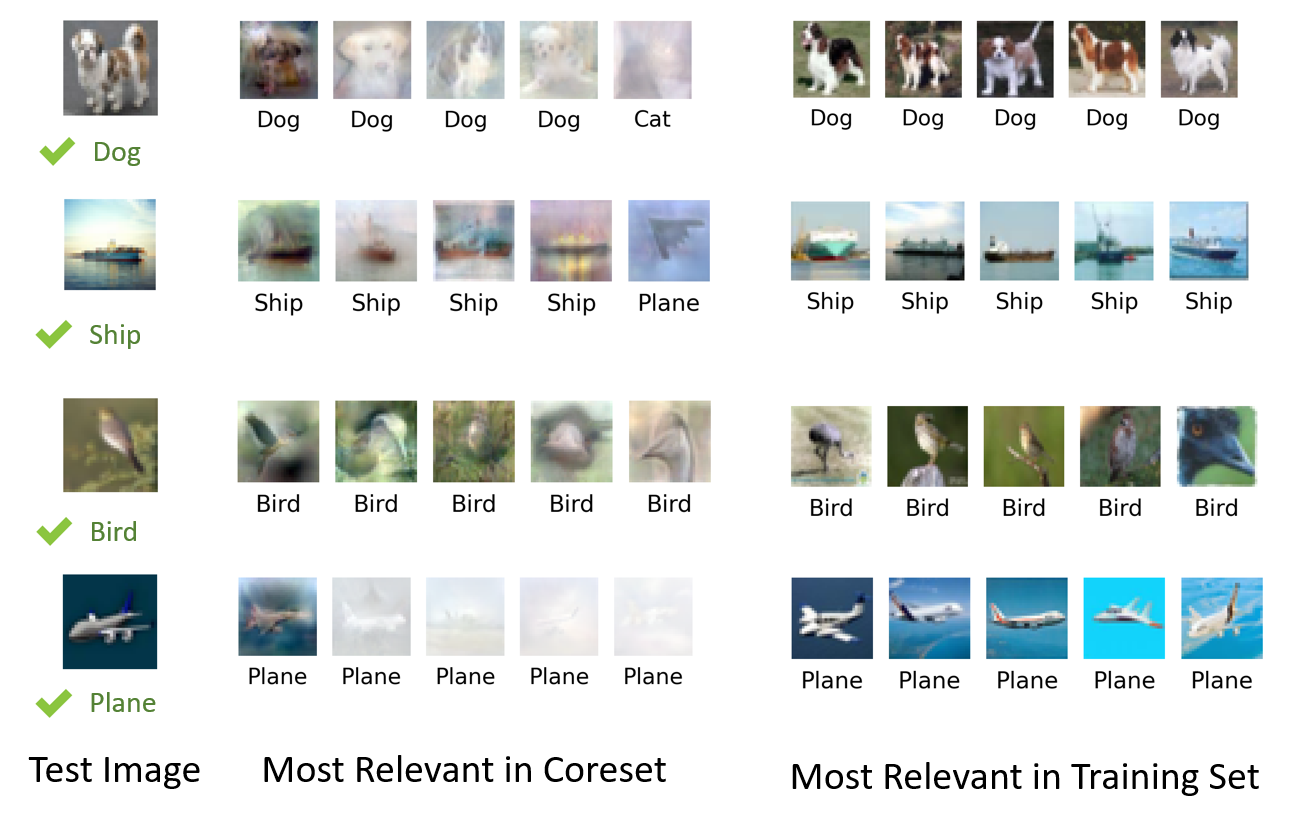}}
\caption{Correctly predicted test images and the most relevant images in the coreset and training set for CIFAR-10, 50 images/cls}
\end{center}
\vskip -0.2in
\end{figure}

\section{Empirical NNGP at Inference additional results}
\label{app:sec_empirical_inference}
\begin{figure}[ht]
\vskip 0.2in
\begin{center}
\centerline{\includegraphics[width=0.8\textwidth]{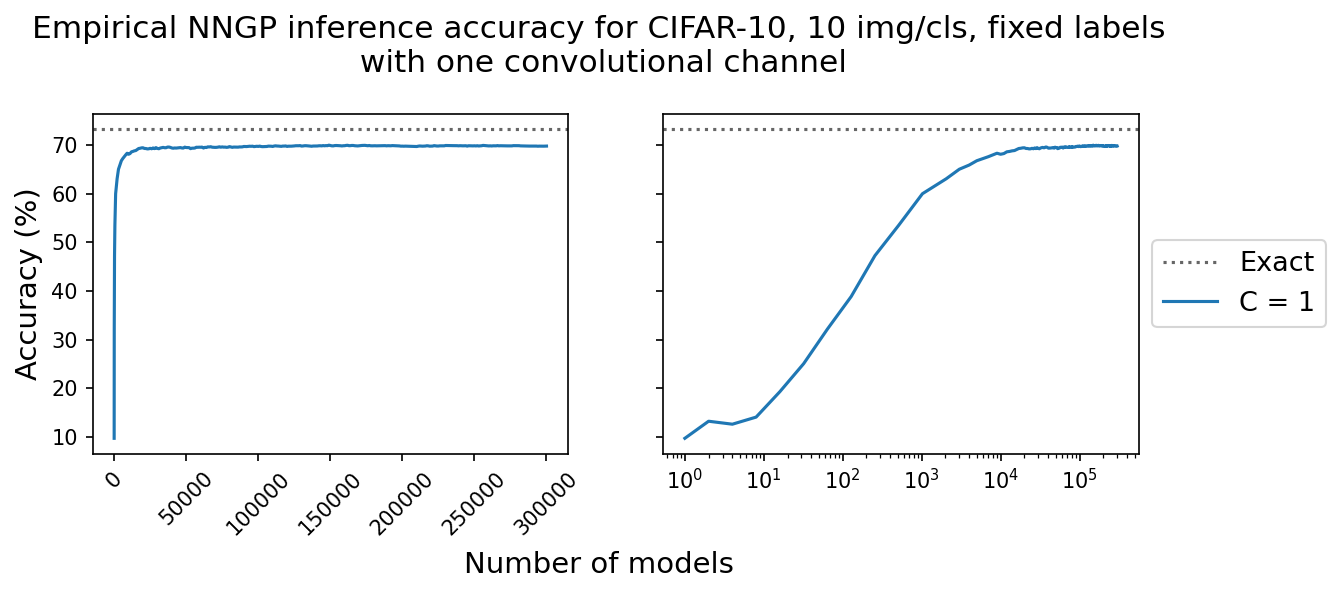}}
\caption{Empirical NNGP performance at inference for CIFAR-10, 10 images per class, fixed labels, using convolutional networks with one convolutional channels. We can achieve reasonable performance, 70\%, albeit requiring over $10^5$ random models.}
\label{fig:app_one_conv}
\end{center}
\vskip -0.2in
\end{figure}

This section contains additional plots showing the effectiveness of using the Empirical NNGP at inference for all of our RFAD distilled datasets. In all cases, we are able to achieve close to the performance of the exact NNGP kernel for convolutional architectures with $C \geq 128$. We additionally show an experiment where we achieve 70\% accuracy on CIFAR 10, using our 10 img/cls fixed label distilled coreset using the empirical NNGP kernel from random neural networks with \textit{one} convolutional channel in \cref{fig:app_one_conv}

\begin{figure}[ht]
\vskip 0.2in
\begin{center}
\subfigure{\includegraphics[width=\columnwidth]{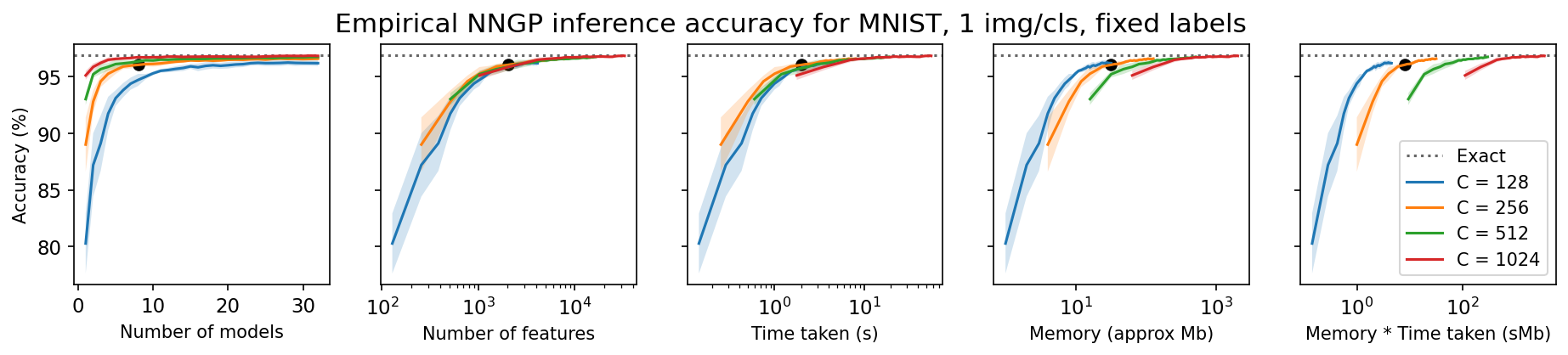}}
\subfigure{\includegraphics[width=\columnwidth]{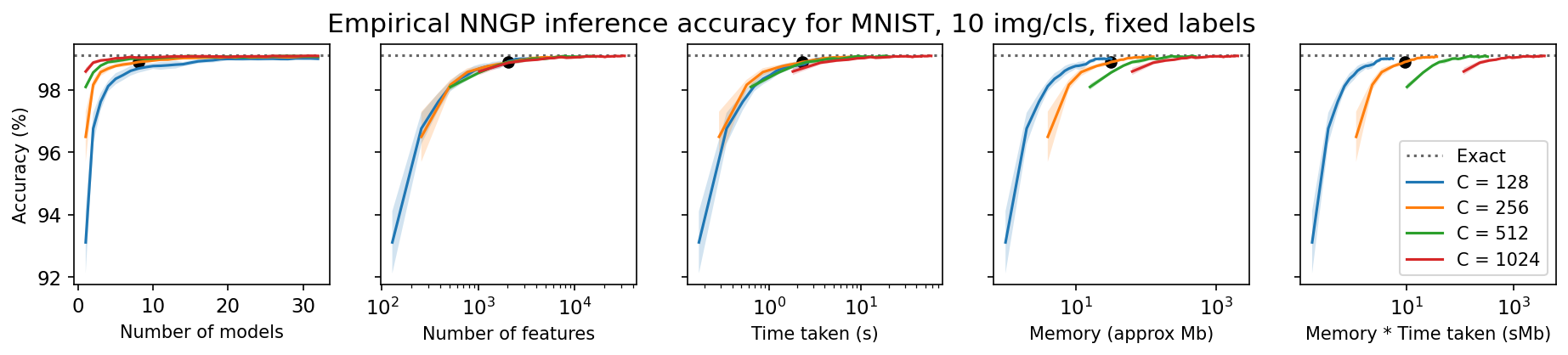}}
\subfigure{\includegraphics[width=\columnwidth]{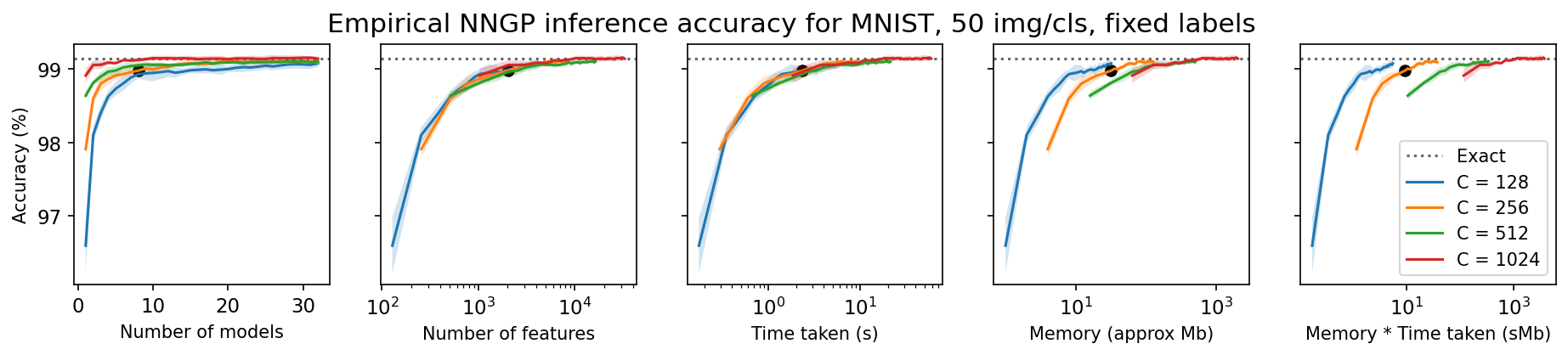}}
\caption{Empirical NNGP inference accuracy for MNIST with fixed labels}
\label{icml-historical2}
\end{center}
\vskip -0.2in
\end{figure}

\begin{figure}[ht]
\vskip 0.2in
\begin{center}
\subfigure{\includegraphics[width=\columnwidth]{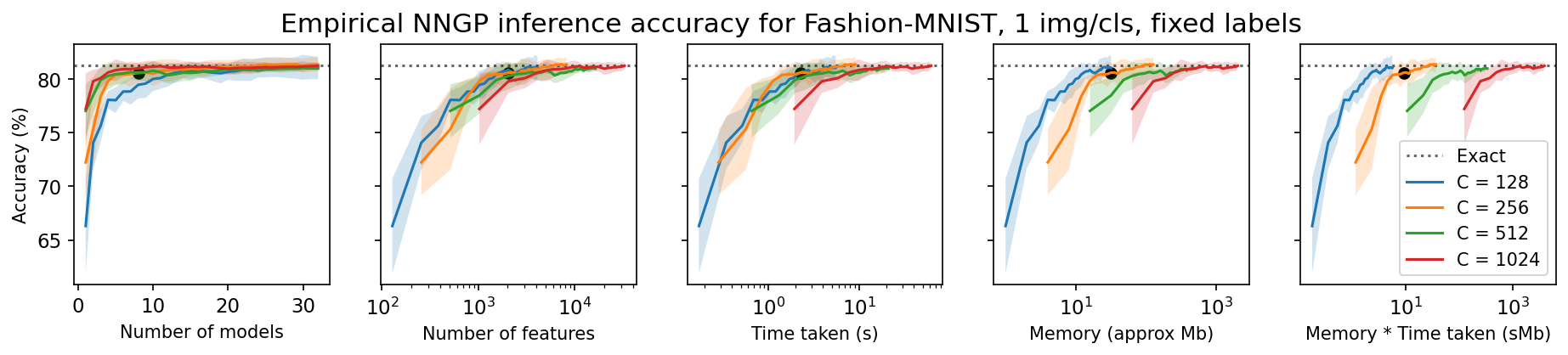}}
\subfigure{\includegraphics[width=\columnwidth]{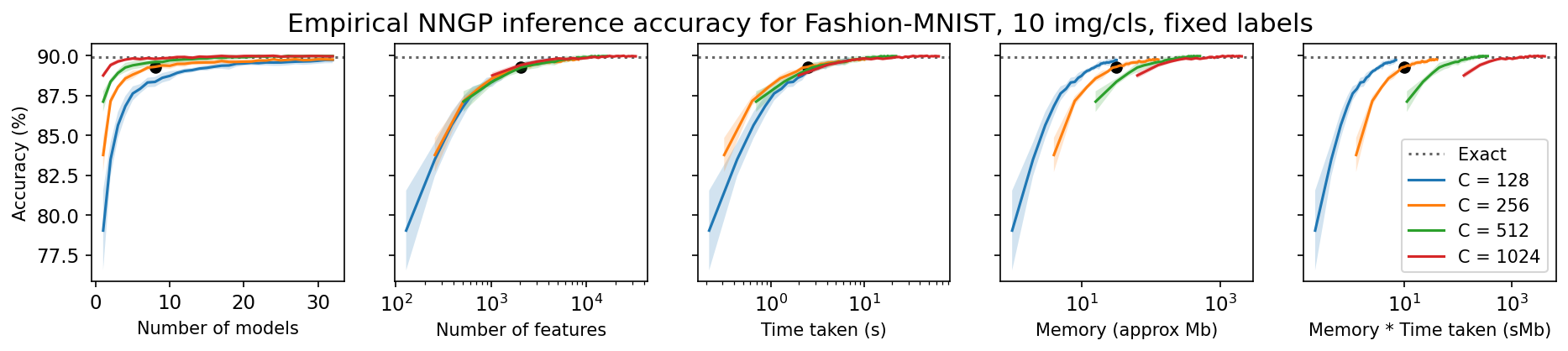}}
\subfigure{\includegraphics[width=\columnwidth]{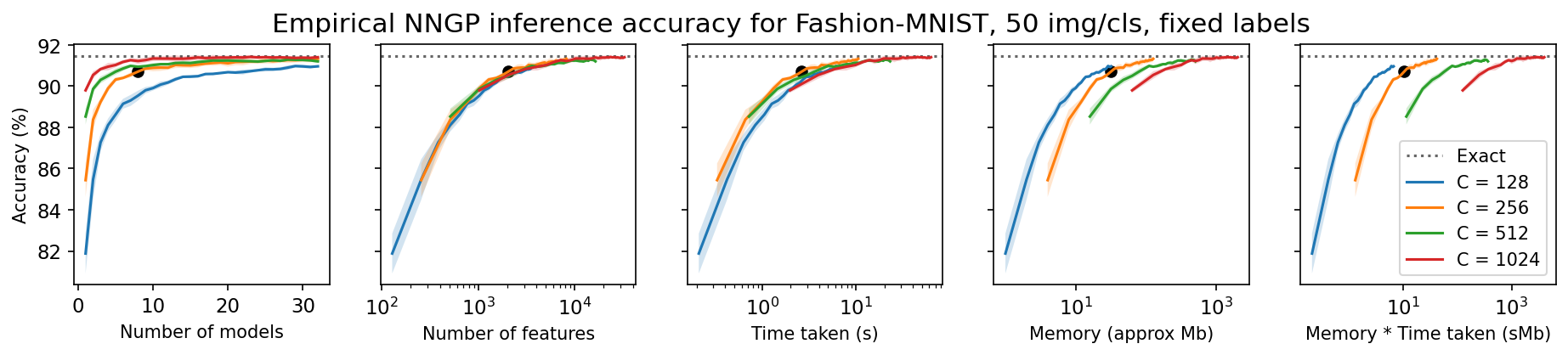}}
\caption{Empirical NNGP inference accuracy for Fashion-MNIST with fixed labels}
\label{icml-historical3}
\end{center}
\vskip -0.2in
\end{figure}

\begin{figure}[ht]
\vskip 0.2in
\begin{center}
\subfigure{\includegraphics[width=\columnwidth]{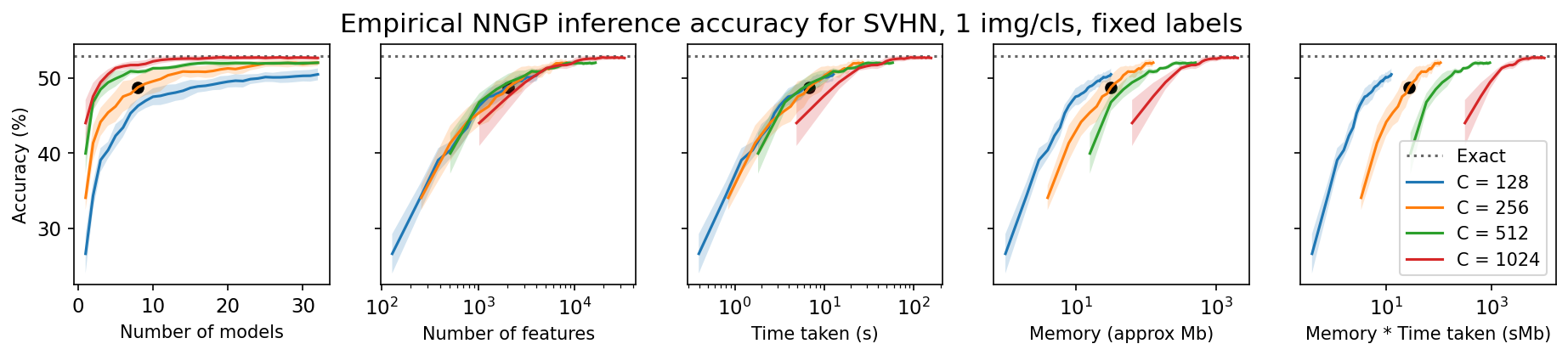}}
\subfigure{\includegraphics[width=\columnwidth]{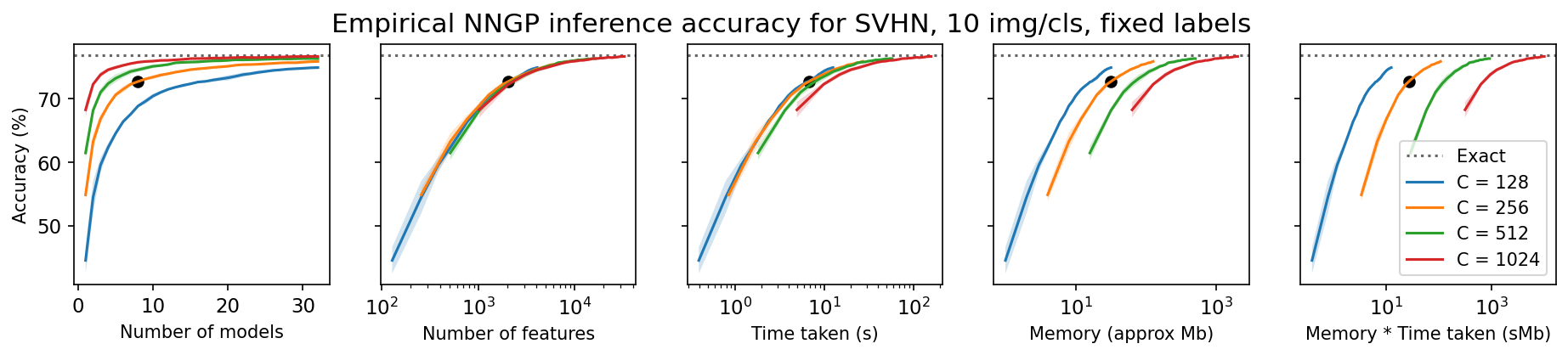}}
\subfigure{\includegraphics[width=\columnwidth]{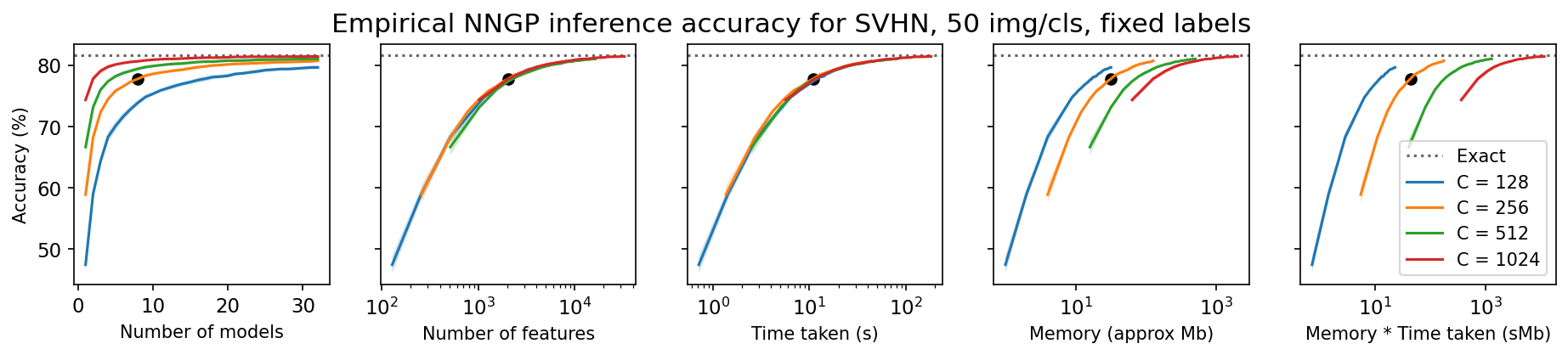}}
\caption{Empirical NNGP inference accuracy for SVHN with fixed labels}
\label{icml-historical4}
\end{center}
\vskip -0.2in
\end{figure}

\begin{figure}[ht]
\vskip 0.2in
\begin{center}
\subfigure{\includegraphics[width=\columnwidth]{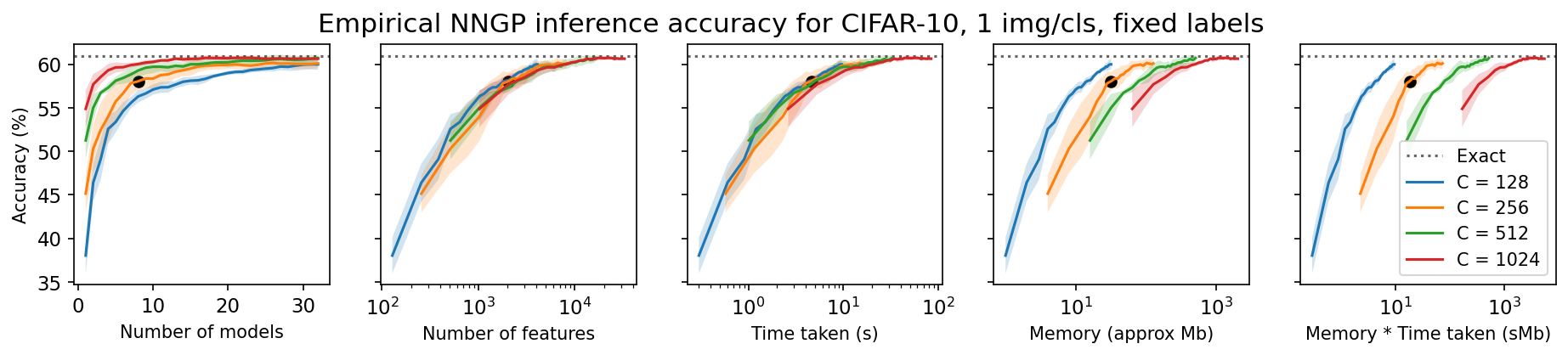}}
\subfigure{\includegraphics[width=\columnwidth]{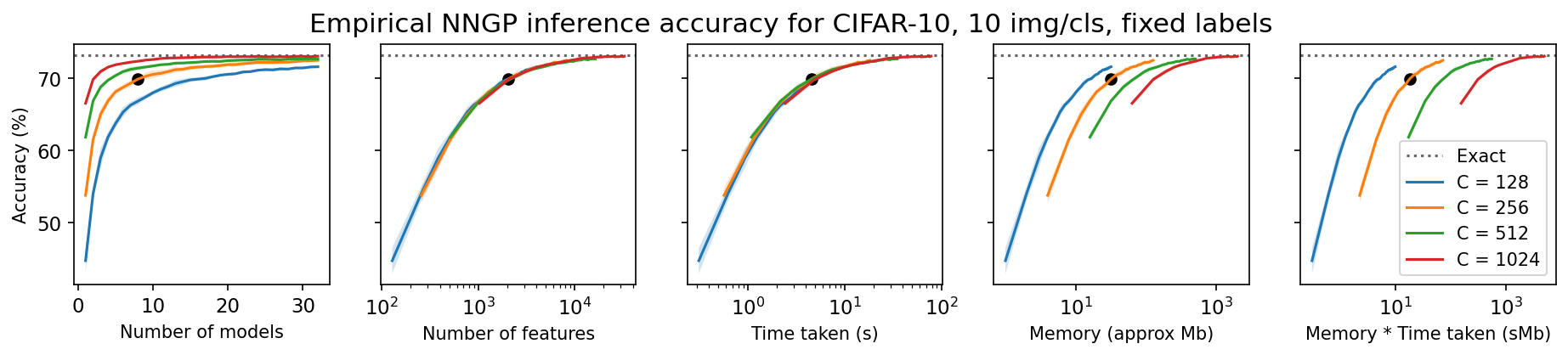}}
\subfigure{\includegraphics[width=\columnwidth]{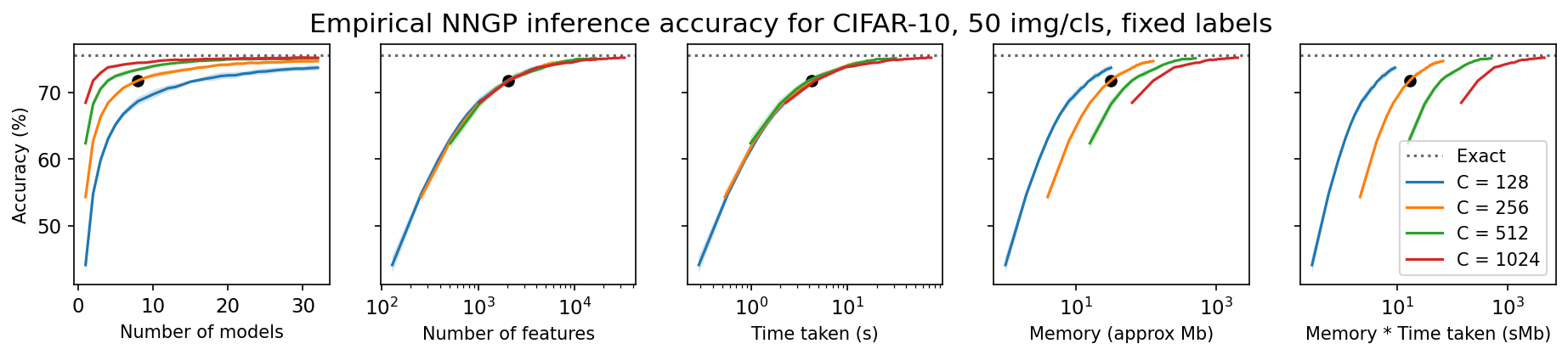}}
\caption{Empirical NNGP inference accuracy for CIFAR-10 with fixed labels}
\label{icml-historical5}
\end{center}
\vskip -0.2in
\end{figure}

\begin{figure}[ht]
\vskip 0.2in
\begin{center}
\subfigure{\includegraphics[width=\columnwidth]{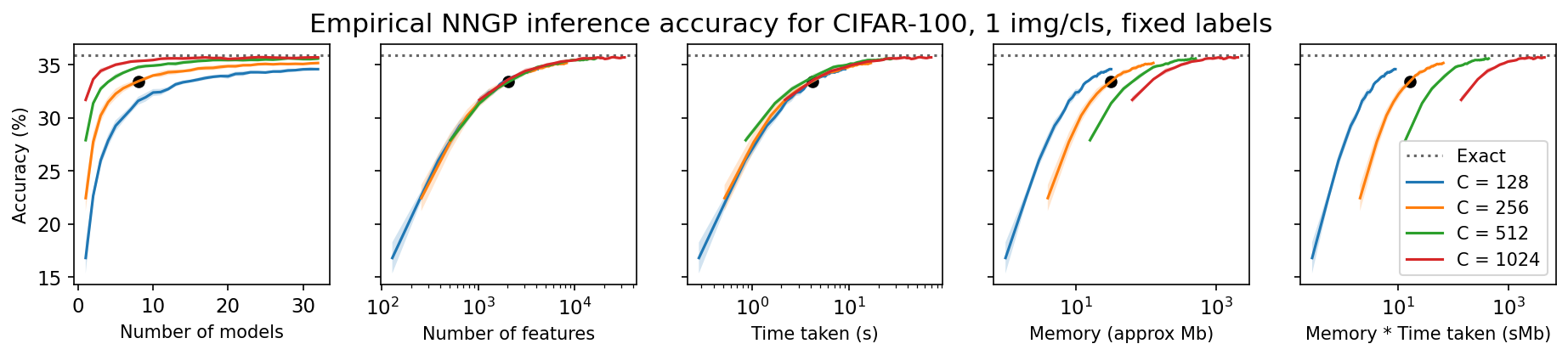}}
\subfigure{\includegraphics[width=\columnwidth]{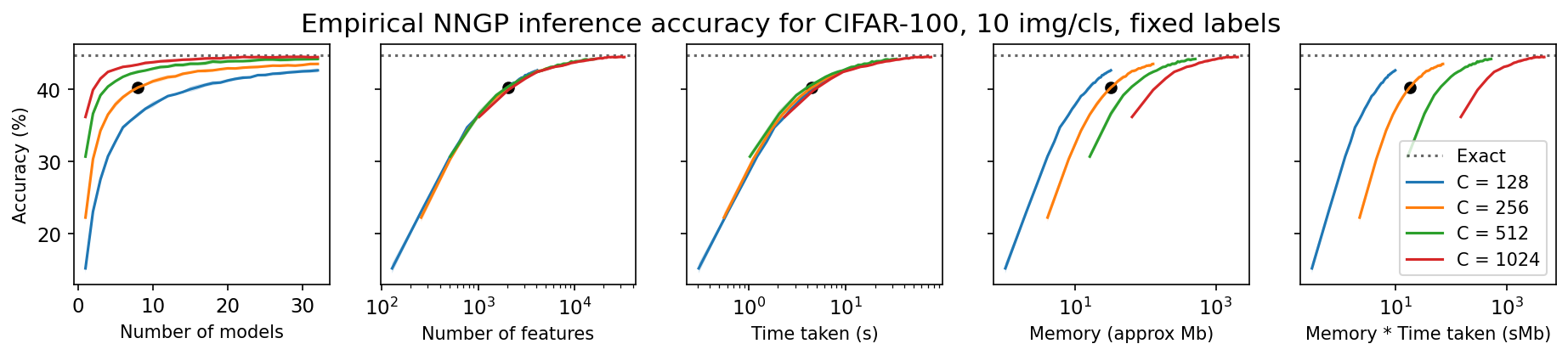}}
\caption{Emperical NNGP inference accuracy for CIFAR-100 with fixed labels}
\label{icml-historical6}
\end{center}
\vskip -0.2in
\end{figure}

\begin{figure}[ht]
\vskip 0.2in
\begin{center}
\subfigure{\includegraphics[width=\columnwidth]{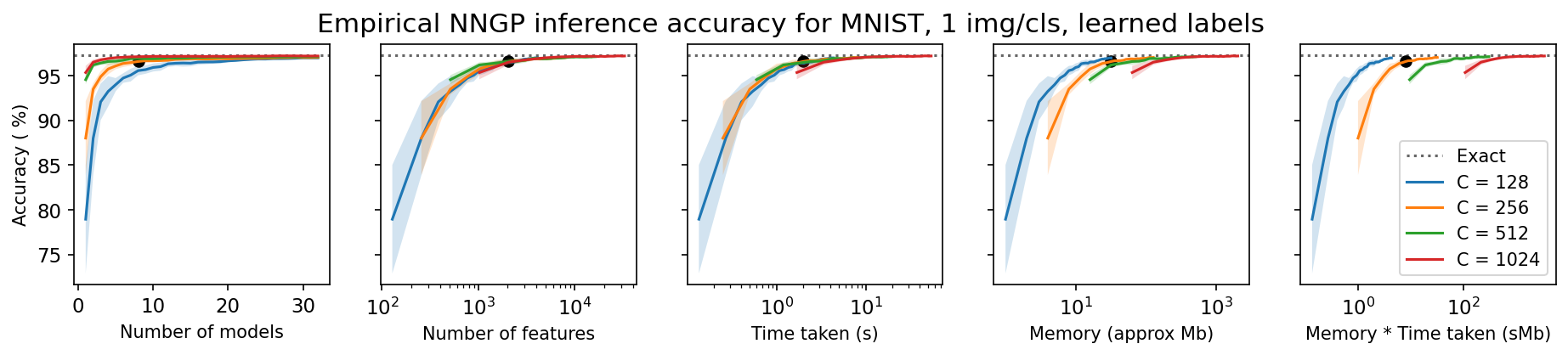}}
\subfigure{\includegraphics[width=\columnwidth]{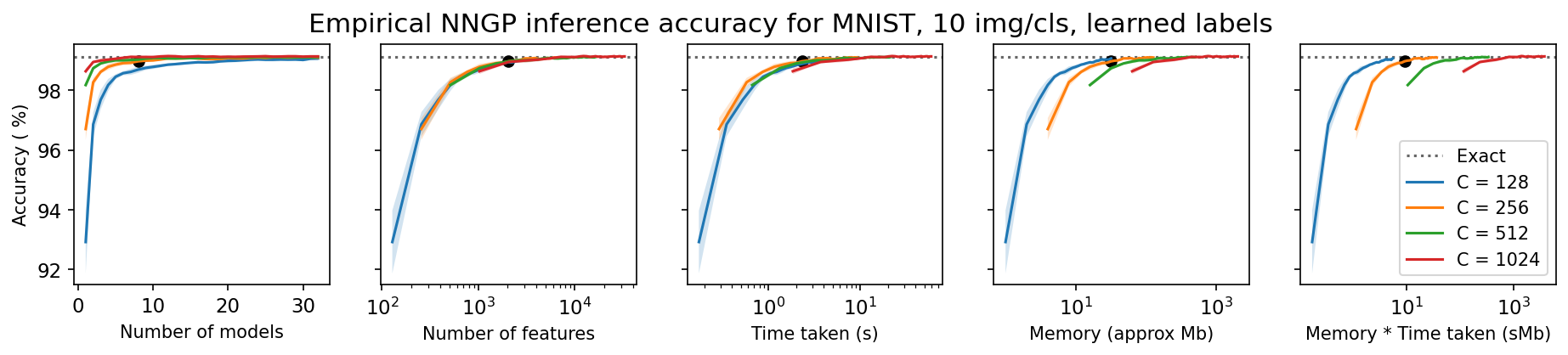}}
\subfigure{\includegraphics[width=\columnwidth]{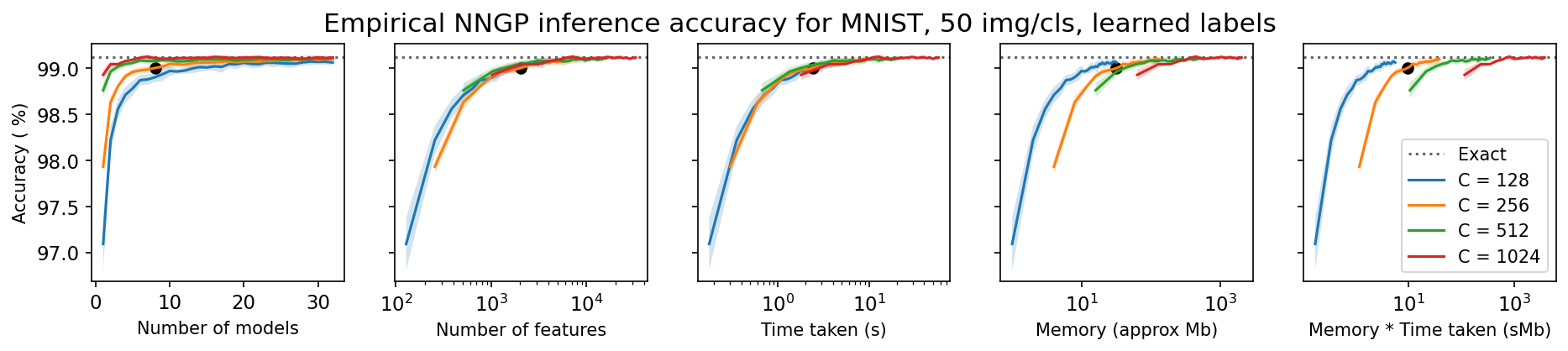}}
\caption{Empirical NNGP inference accuracy for MNIST with learned labels}
\label{icml-historical7}
\end{center}
\vskip -0.2in
\end{figure}

\begin{figure}[ht]
\vskip 0.2in
\begin{center}
\subfigure{\includegraphics[width=\columnwidth]{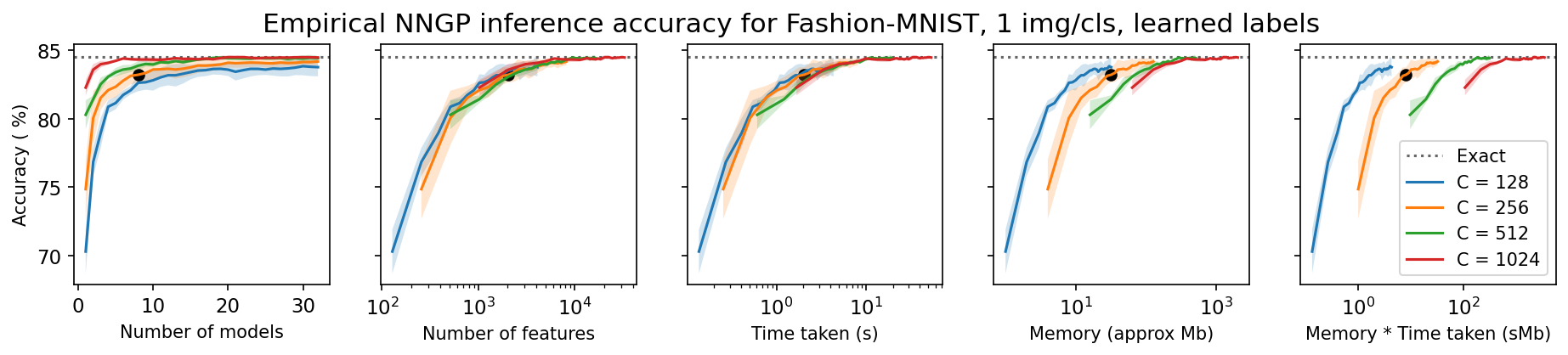}}
\subfigure{\includegraphics[width=\columnwidth]{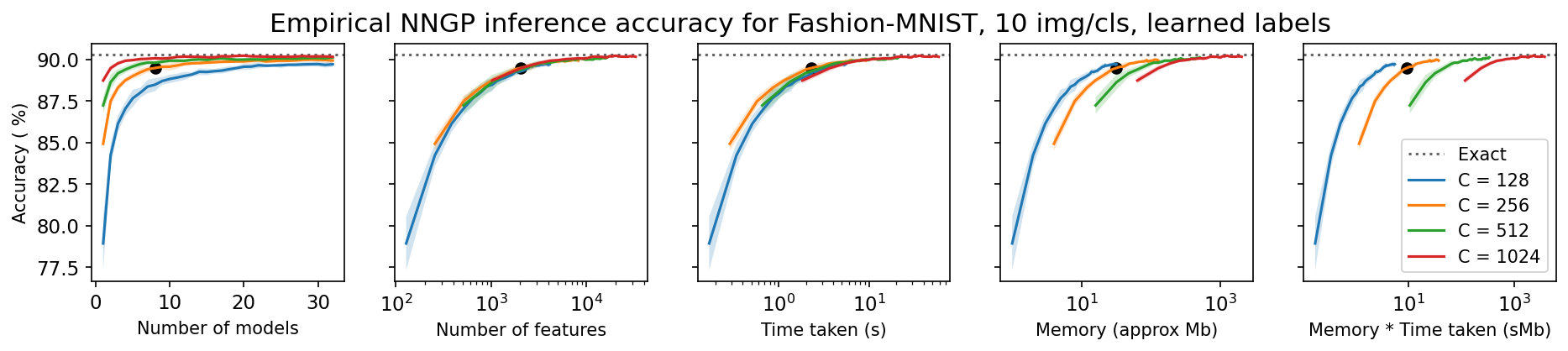}}
\subfigure{\includegraphics[width=\columnwidth]{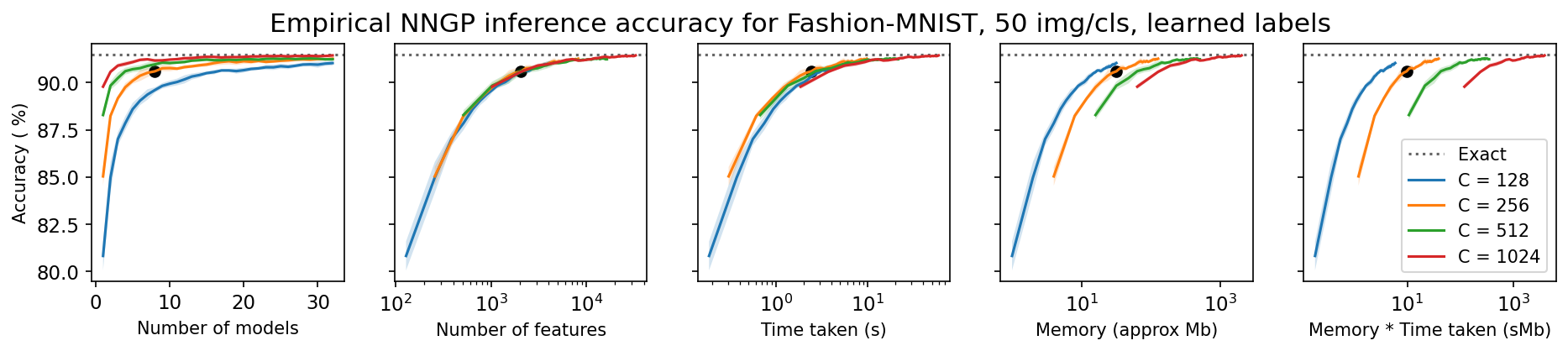}}
\caption{Empirical NNGP inference accuracy for Fashion-MNIST with learned labels}
\label{icml-historical8}
\end{center}
\vskip -0.2in
\end{figure}

\begin{figure}[ht]
\vskip 0.2in
\begin{center}
\subfigure{\includegraphics[width=\columnwidth]{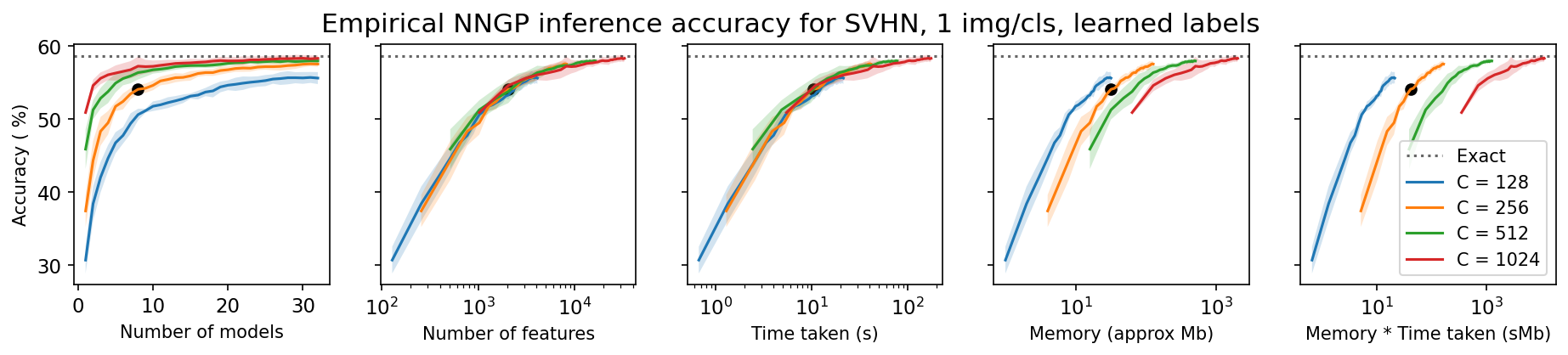}}
\subfigure{\includegraphics[width=\columnwidth]{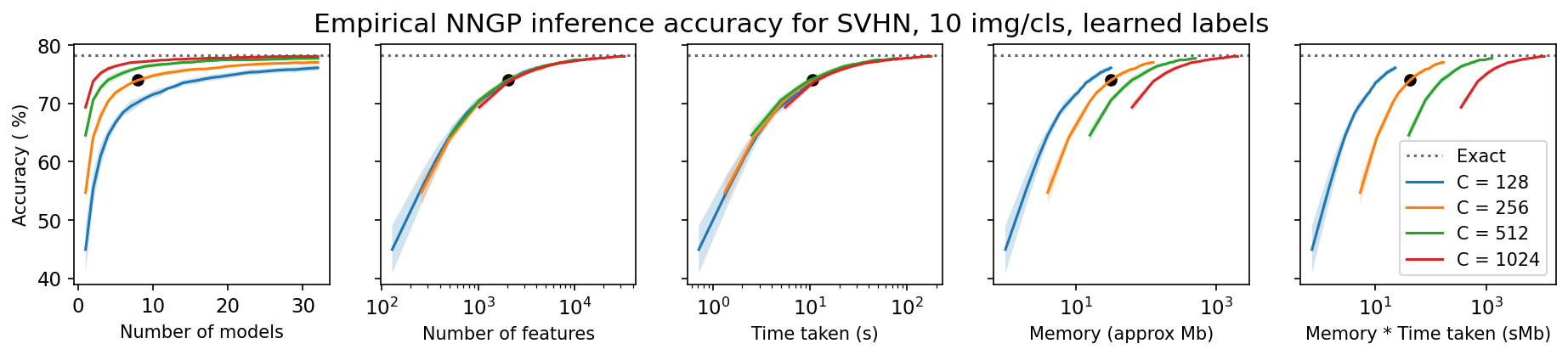}}
\subfigure{\includegraphics[width=\columnwidth]{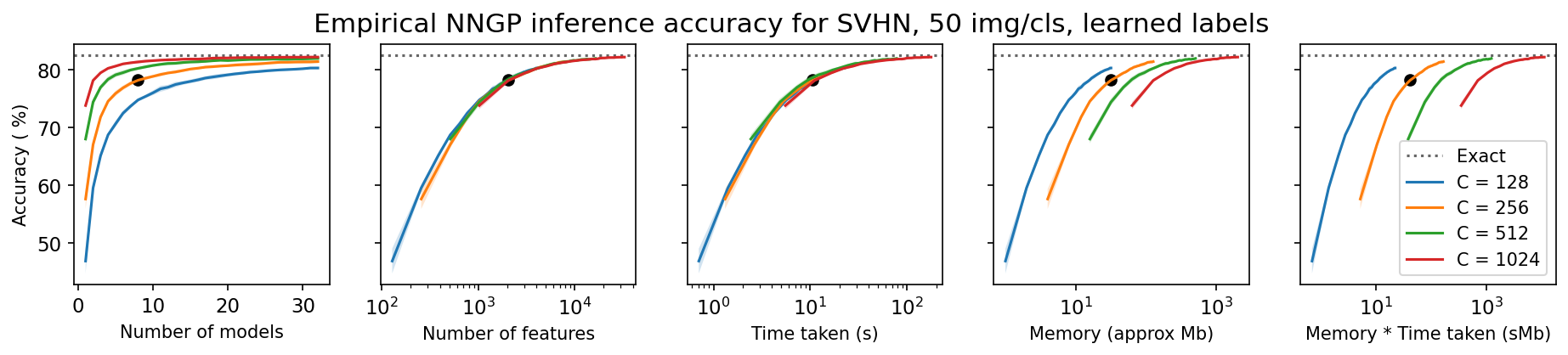}}
\caption{Empirical NNGP inference accuracy for SVHN with learned labels}
\label{icml-historical9}
\end{center}
\vskip -0.2in
\end{figure}

\begin{figure}[ht]
\vskip 0.2in
\begin{center}
\subfigure{\includegraphics[width=\columnwidth]{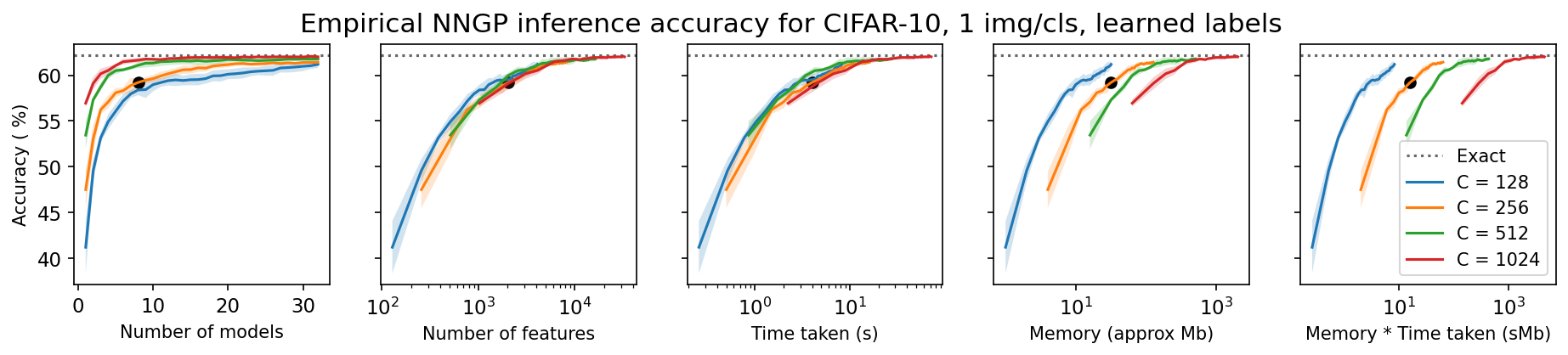}}
\subfigure{\includegraphics[width=\columnwidth]{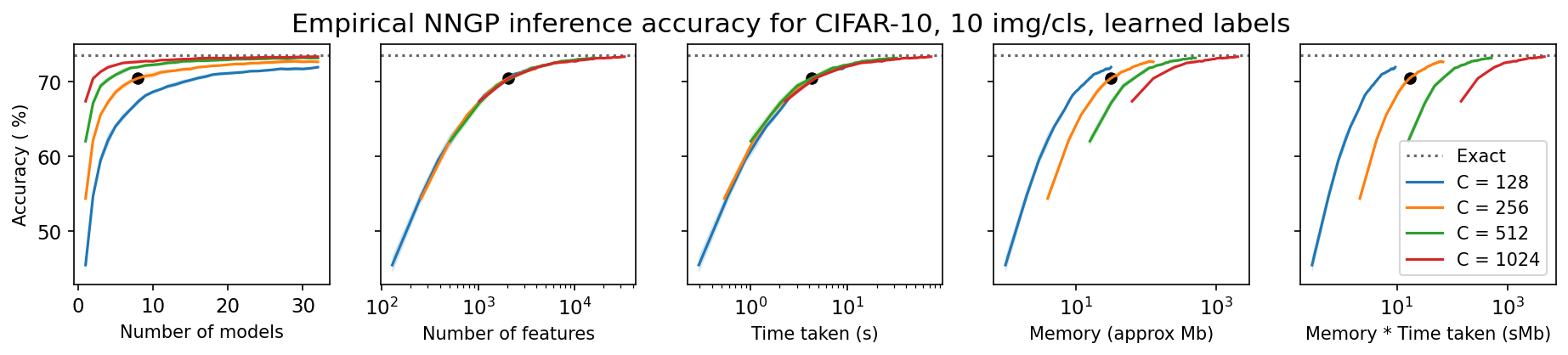}}
\subfigure{\includegraphics[width=\columnwidth]{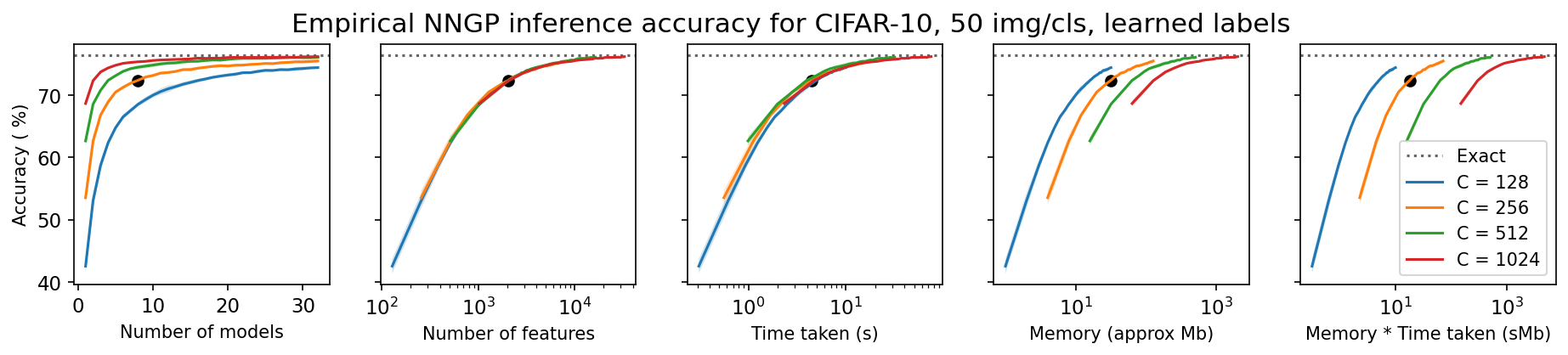}}
\caption{Empirical NNGP inference accuracy for CIFAR-10 with learned labels}
\label{icml-historical}
\end{center}
\vskip -0.2in
\end{figure}

\begin{figure}[ht]
\vskip 0.2in
\begin{center}
\subfigure{\includegraphics[width=\columnwidth]{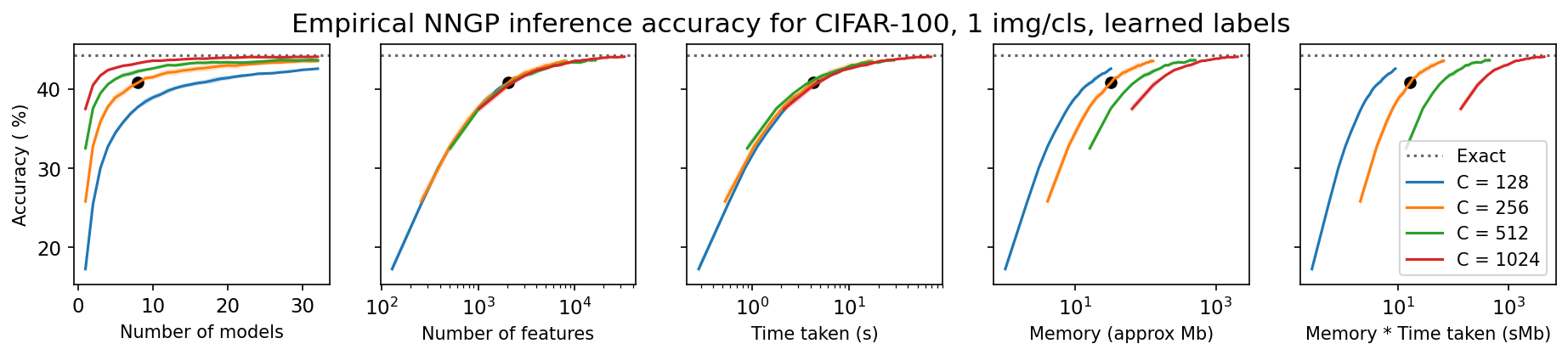}}
\subfigure{\includegraphics[width=\columnwidth]{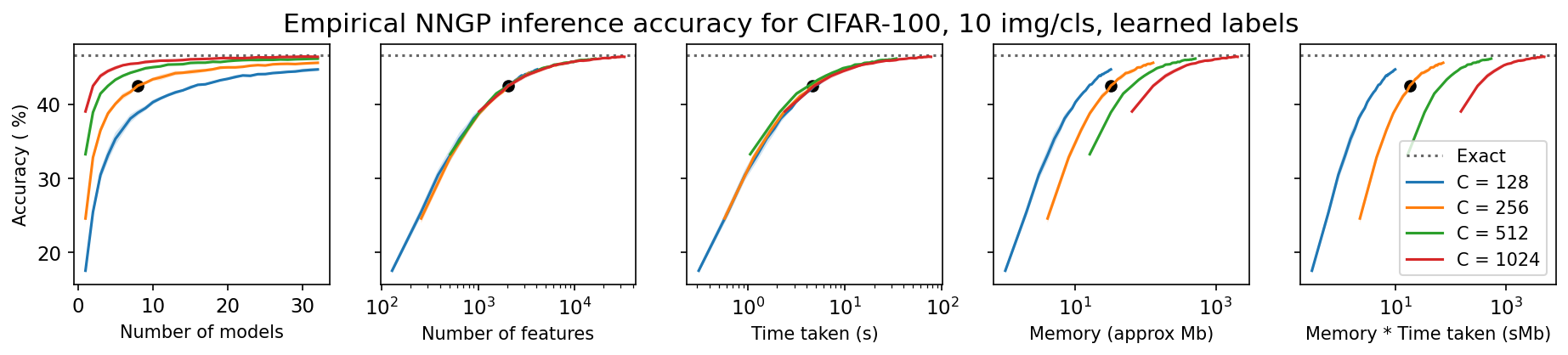}}
\caption{Emperical NNGP inference accuracy for CIFAR-100 with learned labels}
\label{icml-historical10}
\end{center}
\vskip -0.2in
\end{figure}

\end{document}